\documentclass[11pt,a4paper]{article}
\usepackage[hyperref]{emnlp2020}
\usepackage{times}
\usepackage{latexsym}

\usepackage{microtype}

\usepackage{array,multirow}
\usepackage{enumitem}
\usepackage{graphicx}
\usepackage{caption}
\usepackage{subcaption}
\usepackage{amsmath}
\usepackage{amsfonts}
\usepackage{booktabs}
\usepackage{arydshln}
\usepackage{xcolor}
\usepackage{mathtools}
\usepackage{algorithm}
\usepackage{algorithmicx}
\usepackage{algpseudocode}
\graphicspath{ {figs/} }

\aclfinalcopy % Uncomment this line for the final submission
\newcommand{\fmw}{DORB}

\title{DORB: Dynamically Optimizing Multiple Rewards with Bandits}

\author{Ramakanth Pasunuru\;\;\;\;\;\;\; Han Guo\;\;\;\;\;\;\; Mohit Bansal \\
  UNC Chapel Hill \\
  {\tt \{ram, hanguo, mbansal\}@cs.unc.edu} \\
 }

\date{}

\begin{document}
\maketitle
\begin{abstract}
Policy gradients-based reinforcement learning has proven to be a promising approach for directly optimizing non-differentiable evaluation metrics for language generation tasks. However, optimizing for a specific metric reward leads to improvements in mostly that metric only, suggesting that the model is gaming the formulation of that metric in a particular way without often achieving real qualitative improvements. Hence, it is more beneficial to make the model optimize multiple diverse metric rewards jointly. While appealing, this is challenging because one needs to manually decide the importance and scaling weights of these metric rewards. Further, it is important to consider using a dynamic combination and curriculum of metric rewards that flexibly changes over time. Considering the above aspects, in our work, we automate the optimization of multiple metric rewards simultaneously via a multi-armed bandit approach (\fmw{}), where at each round, the bandit chooses which metric reward to optimize next, based on expected arm gains. We use the Exp3 algorithm for bandits and formulate two approaches for bandit rewards: (1) Single Multi-reward Bandit (SM-Bandit); (2) Hierarchical Multi-reward Bandit (HM-Bandit). We empirically show the effectiveness of our approaches via various automatic metrics and human evaluation on two important NLG tasks: question generation and data-to-text generation, including on an unseen-test transfer setup. Finally, we present interpretable analyses of the learned bandit curriculum over the optimized rewards.
\end{abstract}

%%%%%%%%%%%%%%%%%%%%%%%%Introduction Section%%%%%%%%%%%%%%%%%%%%%%%%%%%%%%%%
%%%%%%%%%%%%%%%%%%%%%%%%%%%%%%%%%%%%%%%%%%%%%%%%%%%%%%%%%%%%%%%%%%%%
\section{Introduction}
Recent advancements in end-to-end neural networks-based approaches have shown wide success in various sequence generation tasks: machine translation~\cite{sutskever2014sequence,luong2015effective}, dialogue systems~\cite{vinyals2015neural,serban2016building}, textual summarization~\cite{rush2015neural,nallapati2016abstractive,see2017get}, image/video captioning~\cite{bahdanau2014neural,venugopalan2015sequence,pasunuru2017multi}, question generation~\cite{du2017learning,du2018harvesting,zhang2019addressing}, etc. In all of these tasks, cross-entropy loss optimization has been widely used as a standard optimization approach~\cite{sutskever2014sequence}, but this approach suffers from exposure-bias issue~\cite{ranzato2015sequence} and does not optimize for the non-differentiable automatic evaluation metrics that measure the quality of the generated sequence. Recent introduction of policy gradient-based reinforcement learning approaches address these issues for sequence generation tasks by directly optimizing the non-differentiable evaluation metrics~\cite{zaremba2015reinforcement,ranzato2015sequence,rennie2016self}.

However, optimizing for a particular metric/reward via policy gradient-based approaches often leads to improvement in mostly that specific metric, suggesting that this approach is gaming the metrics~\cite{paulus2017deep}.
The weighted average of multiple metrics or surrogate rewards have been explored~\cite{liu2017improved}, but these approaches have to deal with finding the optimal scale balance across different metrics. One can alternatively optimize multiple metrics via a mixing ratio~\cite{pasunuru2018multi}, but this still needs careful tuning of the mixing ratio. Moreover, all these reward approaches are fixed and do not change over training, and all the metrics may not be important over every stage of the training.
Thus, it might be useful to consider using a dynamic combination of metrics, which rewards to use early vs. later, or which rewards might be useful to come back later in training, and consider the context of the full history of rewards, as well as the model’s current state and the nature of the metric. 

To this end, we present a multi-armed bandit approach (which we name the \fmw{} framework) where the arms of the bandit are the choices of the metrics that we want to optimize as rewards. At every round, the bandit chooses the next possible metric to optimize based on its previous performance history over these metrics, hence allowing the automatic learning of an optimal curriculum of rewards. We explore this approach in the context of exploration vs. exploitation via Exp3 algorithm~\cite{auer2002nonstochastic} with two novel approaches for bandit rewards: (1) Single Multi-reward Bandit (SM-Bandit); (2) Hierarchical Multi-reward Bandit (HM-Bandit). First, we present a reward scaling approach to maintain the metric rewards range in $[0,1]$. Next, we present our SM-Bandit, where at each round, the bandit's reward is based on the performance improvement from multiple sources. Here, we use the average of all the scaled metric rewards from multiple sources as the final reward to the bandit. Finally, we present our HM-Bandit, which consists of a single first-level controller, as well as $K$ second-level multi-armed bandits. The first-level controller's goal is to find the under-performing reward metric, while the second-level bandits' goal is to trigger the specific metric optimizer that will lead to a promising improvement in this specific metric.

We validate the effectiveness of our approaches on two important generation tasks: question generation and data-to-text generation (including an unseen-test transfer setup) via both automatic evaluation metrics and human evaluation. For question generation, we present results on the SQuAD QG dataset~\cite{du2017learning}, and for data-to-text NLG, we choose the WebNLG dataset~\cite{gardent2017webnlg}.
We show that our bandit-based approaches perform statistically significantly better (based on human evaluation) than strong single-reward based RL models as well as non-bandits based multi-reward methods such as the multi-task approach of~\newcite{pasunuru2018multi}. We further present various interpretable analyses of our bandit progress and learned rewards curriculum over different bandit approaches.

%%%%%%%%%%%%%%%%%%%%%%%%Related Work Section%%%%%%%%%%%%%%%%%%%%%%%%%%%%%%%%
%%%%%%%%%%%%%%%%%%%%%%%%%%%%%%%%%%%%%%%%%%%%%%%%%%%%%%%%%%%%%%%%%%%%

\section{Related Works}

\noindent\textbf{Policy Gradient and Generative Models: }
Neural sequence to sequence models with cross-entropy optimization, potentially with attention mechanism~\cite{bahdanau2014neural} and pointer-copy mechanism ~\cite{see2017get,gulcehre2016pointing,vinyals2015pointer,merity2017regularizing}, are widely used in language generation tasks such as machine translation~\cite{sutskever2014sequence,luong2015effective}, abstractive summarization~\cite{chopra2016abstractive,nallapati2016abstractive}, question generation~\cite{du2017learning,zhang2019addressing}, video/image captioning~\cite{xu2015show,vinyals2015show,pasunuru2017multi,zhou2018end}, as well as sentence simplification~\cite{zhang2017sentence,guo2018dynamic}.
However, often the final metrics of interest are not differentiable, and thus not compatible with the standard maximum-likelihood based training.
Motivated by this, recently there has been a surge in applications of reinforcement learning techniques to language generation~\cite{ranzato2015sequence}, in which the gradients of non-differentiable metrics are approximated using the scoring function (REINFORCE~\cite{williams1992simple}). A few successful examples include image captioning~\cite{rennie2016self,ren2017deep}, abstractive summarization~\cite{paulus2017deep,chen2018fast,pasunuru2018multi,celikyilmaz2018deep}, machine translation~\cite{wu2016google,gu2017trainable}, sentence simplification~\cite{zhang2017sentence}, as well as video captioning~\cite{pasunuru2017reinforced,wang2018video}. 
Previous works have explored the problem of optimizing multiple rewards in the context of machine translation~\cite{neubig2016optimization}. For example, the works of~\citet{duh2012learning} and \citet{sankaran2013multi} are based on the theory of Pareto Optimality.
Our approach, instead, dynamically decides the trade-off among metrics, rather than exploring the set of static Pareto-optimal hypotheses.
The most related work on this line is~\citet{pasunuru2018multi}, which simultaneously optimizes multiple rewards in alternate fashion for abstractive summarization. 
In our work, we use a multi-armed bandit framework to dynamically switch among multiple diverse reward optimizations in the context of policy-gradient-based generative models.\footnote{When we say dynamic switching, we mean using one metric at a time (i.e., no explicit weighted combination of loss metrics' optimization for a single mini-batch), but to learn an implicit ratio/proportion of metrics' importance over the overall training trajectory (e.g., BLEU metric might be sampled three times more than ROUGE on average).}

\noindent\textbf{Multi-Armed Bandit:}
Many control problems can be cast as multi-armed bandit problems, where the goal is to select a sequence of arms/actions
in order to optimize certain objective (e.g., expected future payoff)~\cite{bubeck2012regret}. One widely studied problem in the multi-armed bandit literature is finding the optimal trade-off between exploration and exploitation~\cite{audibert2009exploration,macready1998bandit,auer2002finite,kveton2018garbage,bubeck2012regret}.
Some widely used bandit algorithms include $\epsilon$-greedy~\cite{sutton2018reinforcement}, Boltzmann exploration~\cite{kaelbling1996reinforcement}, UCB~\cite{auer2002finite}, Thompson sampling~\cite{chapelle2011empirical}, contextual bandit~\cite{sharaf2019meta}, as well as Exp3 adversarial bandit~\cite{auer2002nonstochastic}. In this work, we use Exp3, and the hierarchical version of it, for the problem of optimizing multiple rewards.\footnote{In our initial experiments, we experimented with a few other bandit approaches (UCB, contextual bandit, variants of Exp3, e.g., Exp3-S), but we ended up with our current Exp3 setting due to its performance and stability reasons within the scope of our methods and tasks.}

Multi-armed bandit algorithms have been used in a wide range of applications, such as online advertising~\cite{chen2013combinatorial}, recommendation~\cite{li2010contextual}, multi-task task selection~\cite{guo2019autosem}, and hyper-parameter optimization~\cite{li2016hyperband,merentitis2019bandit}. 
Recently,~\citet{graves2017automated} apply a non-stationary multi-armed bandit (in particular, the Exp3.S algorithm) to select an adaptive policy (curriculum) that a neural network follows to maximize the learning efficiency.~\citet{sharma2017online} use multi-armed bandit sampling to choose which domain data (harder vs. easier) to feed as input to a single model (using different Atari games). To our knowledge, we are the first ones to apply a multi-armed bandit to optimize multiple rewards in the context of text generation.

%%%%%%%%%%%%%%%%%%%%%%%%Model Section%%%%%%%%%%%%%%%%%%%%%%%%%%%%%%%%%%%%%%%%
%%%%%%%%%%%%%%%%%%%%%%%%%%%%%%%%%%%%%%%%%%%%%%%%%%%%%%%%%%%%%%%%%%%%
\begin{figure*}[t]
\centering
\includegraphics[width=0.93\linewidth]{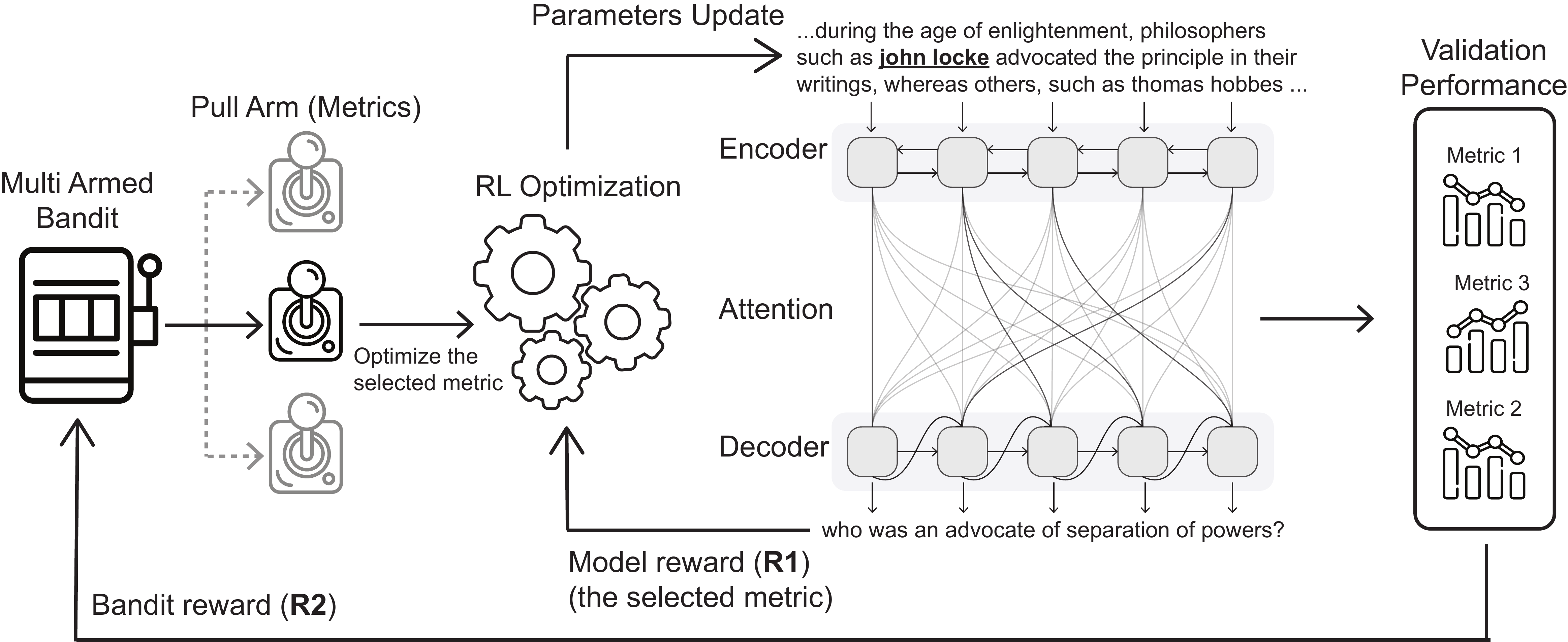}
\vspace{-6pt}
\caption{Overview of our multi-armed bandit reward selection framework \fmw{}. At each step, the model outputs are scored based on a reward function (metric), where the choice of the reward function is dynamically controlled by the multi-armed bandit. Then the corresponding optimization is executed based on the chosen reward function. Finally, the observed validation performance metrics are given as feedback to the bandit.
\label{fig:baseline}
\vspace{-10pt}
}
\end{figure*}

\section{Multi-Reward Optimization}
In this section, we first describe the policy gradients-based reinforcement learning (RL) approach for text generation tasks, and then discuss the need for a better multi-reward optimization approach for RL in the context of generation tasks. Lastly, we introduce our novel methods for multi-reward optimization via multi-armed bandits. 

\noindent\textbf{Glossary:} Agent: RL policy gradients; Bandit: multi-armed bandit; Controller: controller in HM-Bandit (see Fig.~\ref{fig:hierarchical}).

\paragraph{Policy Gradient Background.}
\label{subsec:reinforcement-learning}
Cross-entropy loss based optimization is traditionally used for the sequence generation tasks. However, recent policy gradient-based reinforcement learning approach has shown two advantages over the cross-entropy loss optimization approach: (1) avoiding \emph{exposure bias} issue which is about the mismatch in the output distributions created by different train and test time decoding approaches in cross-entropy loss optimization; (2) able to directly optimize the non-differentiable evaluation metrics. 

To this end, REINFORCE algorithm~\cite{williams1992simple,zaremba2015reinforcement} is used to learn a policy $p_{\theta}$ defined by the model parameters $\theta$ to predict the next action (tokens in our setup). Specifically, instead of minimizing the negative log-likelihood, we minimize the following loss:
\begin{equation}
    L_{\textrm{RL}} = -\mathbb{E}_{w^s \sim p_\theta} [r(w^s)]
\end{equation}
\label{eq:rl}
where $w^s$ is the sequence of sampled tokens and $r(\cdot)$ is the reward function that measures the quality of $w^s$. The derivative of this loss function can then be approximated using a single sample along with a bias estimator $\hat{b}$ to reduce variance:
\begin{equation}
\nabla_\theta L_{\textrm{RL}} = -(r(w^s)-\hat{b}) \nabla_\theta \log p_\theta(w^s)
\end{equation}
There are several ways to calculate the baseline estimator, and in this work we use the SCST mechanism~\cite{rennie2016self}.

\paragraph{Need for a better multi-reward optimization.}
\label{sec:multi-reward-optimization}
Often, an RL agent can improve the policy $p_{\theta}$ via multiple reward sources. However, efficient ways of optimizing multiple rewards in a policy gradient-based reinforcement learning setup have been less explored. Previous works have either explored using a weighted combination of multiple rewards~\cite{zhang2017sentence,li2016deep} or alternate fashion of optimizing multiple rewards inspired via multi-task learning setup~\cite{pasunuru2018multi}. However, these approaches have a disadvantage of tuning the weights of the rewards combination or using a static tunable mixing ratio while optimizing in an alternate fashion. To this end, we explore multi-reward optimization via a multi-armed bandit approach~\cite{bubeck2012regret,LS19bandit-book,burtini2015survey}. During the training, the bandit explores/exploits the choice of reward functions in order to improve the overall performance of the model. In the remaining part of this section, we discuss various multi-armed bandit-based models for multi-reward optimization (Sec.~\ref{subsec:mab-for-multireward}), and reward settings (Sec.~\ref{subsec:bandit-reward-settings}). Then, we present the two novel approaches, namely Single Multi-reward Bandit (SM-Bandit, Sec.~\ref{subsec:sm-bandit}) and Hierarchical Multi-reward Bandit (HM-Bandit, Sec.~\ref{subsec:hm-bandit}).
\subsection{Multi-Armed Bandit for Multi-Reward Optimization}
\label{subsec:mab-for-multireward}
Given a set of $K$ candidate actions (arms) $\{a_1, a_2, ..., a_K\}$, the objective of a multi-armed bandit problem is to maximize rewards earned through a sequence of lever pulls (actions). We call this reward as bandit reward. 
We view the problem of optimizing multiple rewards as a sequential design of experiments~\cite{robbins1952some}, where the bandit's goal is to decide the next arm (loss function) to pull after each round in order to maximize the rewards it earns.

Let $\{R_1, R_2,..,R_K\}$ be a set of different rewards from $K$ sources which can measure the model/policy's performance. To directly maximize the performance of these $K$ rewards, we need to use $K$ different reinforcement learning-based loss functions. Let the loss function for $R_i$ be:
\begin{equation}
    L_{\operatorname{RL}_i} = -\mathbb{E}_{w^s \sim p_\theta} [R_i(w^s)]
\end{equation}
Each of these $K$ loss functions is considered as an arm of the multi-armed bandit (i.e., the arms/joysticks in Fig.~\ref{fig:baseline}), where pulling the $i^{th}$ arm will result in optimizing for reinforcement based loss function $L_{\operatorname{RL}_i}$ (i.e., in Fig.~\ref{fig:baseline}, main model parameters get updated). The goal of the bandit is to explore and exploit different loss functions and maximize its reward (the validation performance of the model, see Fig.~\ref{fig:baseline}). One widely studied problem is the trade-off between ``exploitation'' of the arm with the highest estimated payoff and ``exploration'' of less known arms. For this, we use the popular Exp3 bandit algorithm~\cite{auer2002nonstochastic} (see Appendix~\ref{sec:exp3-bandit-algo} for more details on Exp3).

\subsection{Bandit Reward Settings}
\label{subsec:bandit-reward-settings}
\label{subsec:bandit-reward-setting}
Note that in this work, we have two sets of rewards: rewards used for optimizing the sequence generation model via policy gradients-based reinforcement learning (R1 in Fig.~\ref{fig:baseline}, Sec.~\ref{subsec:reinforcement-learning}), and rewards used for the bandit (R2 in Fig.~\ref{fig:baseline}). The rewards for the generation model are used to optimize the model w.r.t. the metric of interest, while the rewards for the bandit help the bandit decide which ``metric of interest'' the generation model should optimize.

In order to maintain consistent magnitude/scale across metric rewards while using them for bandits, we use scaled rewards via the quantiles of rewards history following~\citet{graves2017automated}. Let $\mathbf{R}^t=\{R^i\}_{i=1}^{t-1}$ be the history of unscaled rewards up to time step $t$. Let $q_t^{lo}$ and $q_t^{hi}$ be the lower and upper quantiles of $\mathbf{R}^t$, respectively.\footnote{We set $q_t^{lo}$ and $q_t^{hi}$ to be 20$^{th}$ and 80$^{th}$ quantiles.} Then, the scaled reward, $\hat{r}^t$ is defined as follows:
\begin{equation}
\label{eq:scaled-reward}
    \hat{r}^t = 
   \begin{cases}
    0 & \text{if  } R^t < q_t^{lo} \\
    1 & \text{if  } R^t > q_t^{hi} \\
    \frac{R^t - q_t^{lo}}{q_t^{hi} - q_t^{lo}},              & \text{otherwise}
\end{cases} 
\end{equation}
Instead of keeping the entire history of rewards, we use past $n$ rewards from the history.

\begin{algorithm}[t]
\begin{small}
\caption{SM-Bandit Training}
\label{alg:sm-bandit}
\begin{algorithmic}[1]
\State \textbf{Inputs:} \#rewards: $K$, \#train steps: $n_{\textrm{train}}$, \#steps in bandit round: $n_{\textrm{bandit}}$ 
\State Initialize the Exp3 bandit $B$ with $K$ arms
\State $a \gets$ chooseArm($B$) \Comment{Based on Eqn.~\ref{eq:choosen-arm}}
\State $i \gets 0$
\While{$i < n_{\textrm{train}}$}
\State Sample word sequence $w^s$ from model
\State Calculate rewards $R^{\textrm{train}}$ based on $w^s$
\State Optimize model's $L_{\operatorname{RL}_a}$ loss using $R^{\textrm{train}}_{a}$
\If{$i \mod n_{\textrm{bandit}} == 0 $}
\State Evaluate model to get $R^{\textrm{val}}$
\State $r \gets \frac{1}{K}\sum^K_{k=1} \textrm{scaled}(R^{\textrm{val}}_{k})$  \Comment{Based on Eqn.~\ref{eq:scaled-reward}}
\State updateBandit($B, a, r$)  \Comment{Based on Eqn.~\ref{eq:update-bandit}}
\State $a \gets$ chooseArm($B$)
\EndIf
\State $i \gets i + 1$
\EndWhile
\end{algorithmic}
\end{small}
\end{algorithm}

\subsection{Single Bandit with Multi-Reward}
\label{subsec:sm-bandit}
Often, we want to optimize multiple metrics in our RL approach. For this, we have to give a joint reward coming from multiple sources (metrics in our case) to the bandit as a bandit reward. One can easily give the weighted combination of these rewards coming from multiple sources as a reward to the bandit. However, tuning these weights is intractable if the number of reward sources is large. Here, we introduce a new approach called Single Multi-reward Bandit (SM-Bandit), which avoids tuning and uses rewards from multiple sources as feedback to the bandit. 
Let $L_{\operatorname{RL}_1}$, $L_{\operatorname{RL}_2}$, and $L_{\operatorname{RL}_3}$ be the reinforcement learning-based loss functions corresponding to three arms of the bandit: $\operatorname{arm}_1$, $\operatorname{arm}_2$, and $\operatorname{arm}_3$, respectively. If $\operatorname{arm}_2$ is selected at round $t$, then we optimize for $L_{\operatorname{RL}_2}$ and measure the performance of all the unscaled metric scores on the validation set and then calculate the corresponding scaled rewards for each metric. We average over these scaled rewards and give that as a reward to the bandit. The generalization of this reward for $K$-armed bandit is: $r^t = \frac{1}{K} \sum_{i=1}^K \hat{r}_i^t$, where $r^t$ is the bandit reward at round $t$ and $\hat{r}^t_i$ is the scaled reward (Eq.~\ref{eq:scaled-reward}) for the metric corresponding to $\operatorname{arm}_i$ at round $t$. This approach allows us to avoid tuning the balancing weights across the metrics that we optimize, and ensure that the bandit is improving all metrics, as the bandit goal is to maximize the average of all metrics. A detailed procedure of SM-Bandit is described in Algorithm~\ref{alg:sm-bandit}.

\begin{figure}[t]
\centering
\includegraphics[width=0.77\linewidth]{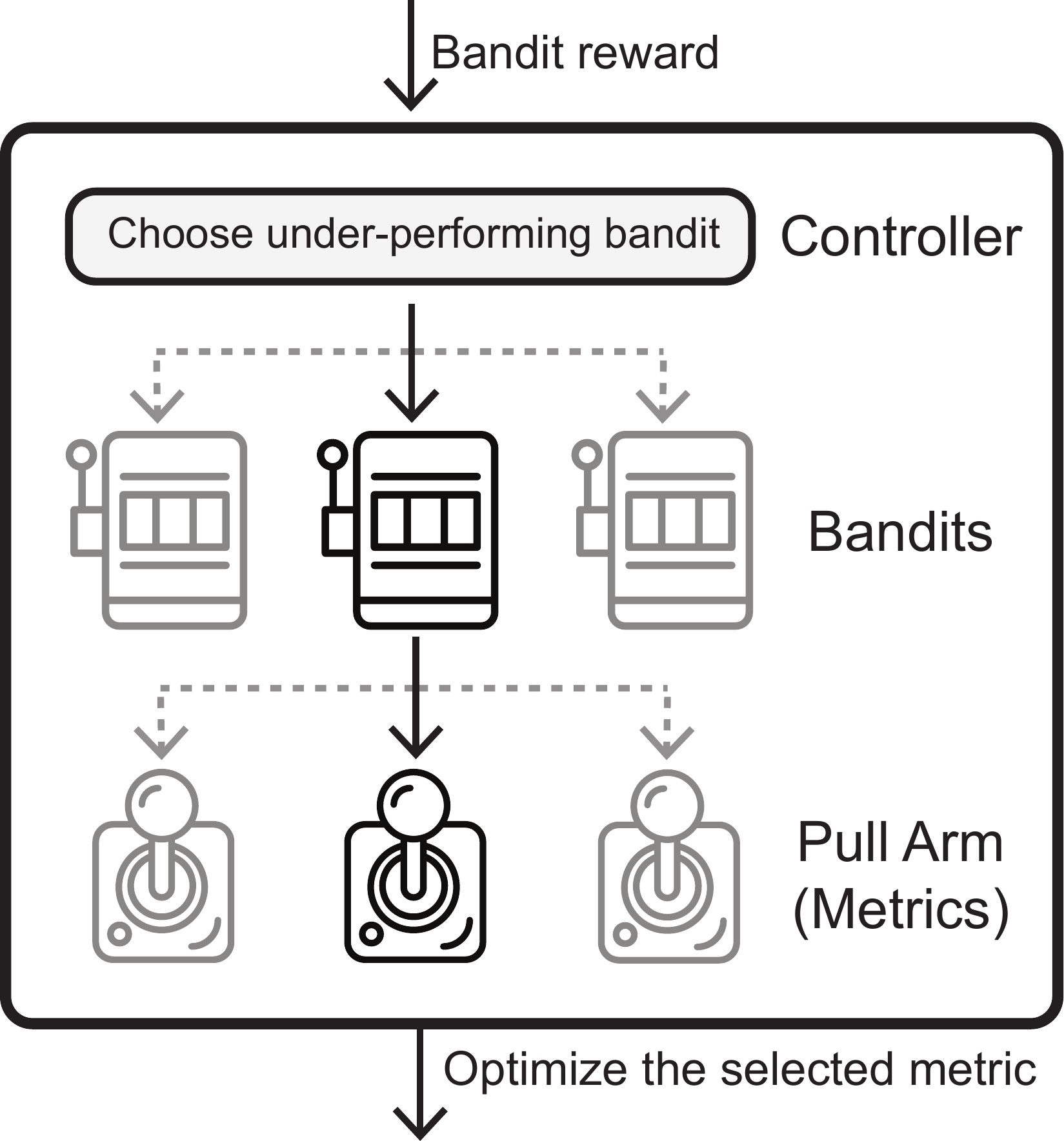}
\vspace{-5pt}
\caption{Overview of the hierarchical multi-armed bandit. The first-level has a controller and the second-level has bandits. The controller decides which bandit of the second-level will be pulled. The second-level bandits then decide which metric to use as the reward function during RL optimization.
\label{fig:hierarchical}
\vspace{-10pt}
}
\end{figure}

\begin{algorithm}[t]
\begin{small}
\caption{HM-Bandit Training}
\label{alg:hm-bandit}
\begin{algorithmic}[1]
\State \textbf{Inputs:} \#rewards: $K$, \#train steps: $n_{\textrm{train}}$, \#steps in bandit round: $n_{\textrm{bandit}}$, \#steps in controller round:  $n_{\textrm{controller}}$ 
\State Create the controller $C$ with $K$ bandits
\State Initialize all bandits, and set $j \gets 0$
\State $B \gets \textrm{chooseBandit}(C, j)$ \Comment{choose bandit at index $j$}
\State $a \gets$ chooseArm($B$) \Comment{Based on Eqn.~\ref{eq:choosen-arm}}
\State $i \gets 0$
\While{$i < n_{\textrm{train}}$}
\State Sample word sequence $w^s$ from model
\State Calculate rewards $R^{\textrm{train}}$ based on $w^s$
\State Optimize model's $L_{\operatorname{RL}_a}$ loss using $R^{\textrm{train}}_{a}$
\If{$i \mod n_{\textrm{bandit}} == 0 $}
\State Evaluate model to get $R^{\textrm{val}}$
\State $r \gets \textrm{scaled}(R^{\textrm{val}}_{j})$
\State updateBandit($B, a, r$) \Comment{Based on Eqn.~\ref{eq:update-bandit}}
\State $a \gets$ chooseArm($B$)
\EndIf
\If{$i \mod n_{\textrm{controller}} == 0 $}
\State Evaluate model to get $R^{\textrm{val}}$
\State $j \gets \operatorname*{argmin}_{k} \{\textrm{scale}(R^{\textrm{val}}_{k})\}_{k=1}^K$
\State $B \gets \textrm{chooseBandit}(C, j)$
\State $a \gets$ chooseArm($B$)
\EndIf
\State $i \gets i + 1$
\EndWhile
\end{algorithmic}
\end{small}
\end{algorithm}

\subsection{Hierarchical Bandit with Multi-Reward}
\label{subsec:hm-bandit}
The SM-bandit's goal in the previous approach described in Sec.~\ref{subsec:sm-bandit} is to improve all metrics using a single bandit.
In this section, we introduce another bandit-based variant to improve all metrics but by using multiple bandits which are controlled by a controller, called Hierarchical Multi-reward Bandits (HM-Bandit, Fig.~\ref{fig:hierarchical}). The HM-Bandit consists of a single first-level controller (not a bandit, top row in Fig.~\ref{fig:hierarchical}), and $K$ second-level multi-armed bandits (middle row in Fig.~\ref{fig:hierarchical}). The first-level controller's goal is to find the under-performing reward metric, while the second-level bandits' goal is to trigger a specific metric optimizer that will lead to a promising improvement in this specific metric. More intuitively, the first-level controller sets the objective (e.g., ROUGE needs to be improved), while the second-level bandit decides which specific reward function can help accomplish the objective. A detailed procedure of our HM-bandit is described in Algorithm~\ref{alg:hm-bandit}. This concept is also loosely related to Bayesian model selection, where it's common to use a hierarchical specification of models~\cite{Rasmussen2015GPML}.

%%%%%%%%%%%%%%%%%%%%%%%%Task Description Section%%%%%%%%%%%%%%%%%%%%%%%%%%%%%%%%%%
%%%%%%%%%%%%%%%%%%%%%%%%%%%%%%%%%%%%%%%%%%%%%%%%%%%%%%%%%%%%%%%%%%%%

% 
% 
% 

\section{Tasks and Setup}
We use question generation and data-to-text generation tasks in our experiments. In this section, we discuss the details on these two tasks along with the experimental setup.
\subsection{Question Generation}
The goal of the question generation (QG) task is to generate a natural question that can be answered by the given answer span in a context. Recent works have applied seq2seq neural models for QG, e.g., generating the question given answer sentence~\cite{du2017learning, zhou2017neural}, or the whole paragraph~\cite{du2018harvesting,song2018leveraging,liu2019learning,zhao2018paragraph,kim2018improving, sun2018answer}. Many works also used RL to optimize specific metrics~\cite{song2017unified,kumar2018framework,yuan2017machine}. Recently,~\citet{zhang2019addressing} proposed semantics-enhanced rewards to improve the QG model, and also used the multi-reward approach proposed by~\newcite{pasunuru2018multi} in their RL models.

\paragraph{Baseline.}
Given a paragraph $p$, and an answer span $a$, the goal of the QG model is to generate a question $q$ answering $a$. We follow the encoder-attention-decoder style architecture (see Fig.~\ref{fig:baseline}). The encoder is a bi-directional LSTM-RNN~\cite{hochreiter1997long} with self-attention~\cite{wang2017gated}, and the decoder is a uni-directional LSTM-RNN with attention~\cite{luong2015effective} and pointer~\cite{gu2016incorporating} mechanism, similar to~\citet{zhang2019addressing}. The input to the model is a concatenation of contextualized word representations (BERT~\cite{devlin2019bert}), answer tag embedding (BIO tagging scheme), Part-of-Speech (POS) tag embedding, and Named-Entity (NER) tag embedding. 

\paragraph{Rewards.}
We use ROUGE-L, QPP, and QAP~\cite{zhang2019addressing} as rewards for this task. QPP is calculated as the probability of the generated question being the paraphrase of the ground-truth question via a classifier trained on Quora Question Pairs. QAP is calculated as the probability of a pre-trained QA model to correctly answer the given generated question as input.

\paragraph{Dataset \& Evaluation.}

We use the SQuAD QG English dataset from~\citet{du2017learning} for the QG task, derived from SQuAD v1.1~\cite{rajpurkar2016squad}, and the test set consists of 10\% sampled examples from the training set, as the SQuAD test set is not open. 
For pre-processing, we do standard tokenization.  We report on evaluation metrics including BLEU-4, METEOR, ROUGE-L, Q-BLEU1~\cite{nema2018towards}, as well as QPP and QAP~\cite{zhang2019addressing}.

\subsection{Data-to-Text Generation}
Data-to-text is the task of expressing the components (attributes and values) of meaning representation (MR) as human-readable natural sentences. Previous work in this area include templates~\cite{reiter1995nlg}, rules~\cite{reiter2005choosing}, pipelines~\cite{reiter2007architecture,reiter1997building}, probabilistic models~\cite{liang2009learning} and more recently end-to-end as well as neural-based methods~\cite{wen2015semantically,mei2015talk,duvsek2016sequence,lampouras2016imitation,dusek2019e2e,wiseman2017challenges,gong2018technical,chen2008learning,reiter2017you,lebret2016neural,distiawan2018gtr,gehrmann2018end,marcheggiani2018deep,guo2019densely,zhao2020bridging}. In our work, we use the state-of-the-art model from~\newcite{zhao2020bridging} as our baseline.

\paragraph{Baseline.}
Given a set of Resource Description Framework (RDF) triples,\footnote{Each triple contains a subject, a predicate, and an object.} the task is to generate a natural language text describing the facts in the RDF data. Following~\newcite{zhao2020bridging}, we serialize and reorder the RDF data as an intermediate planning setup, and feed the plan into a seq2seq model with attention and copy mechanism.

\paragraph{Rewards.}
We use BLEU, ROUGE-L, and Entailment-Score~\cite{pasunuru2018multi} as rewards. Entailment-Score is calculated based on the probability that the generated sentence is classified as an entailment w.r.t. the ground truth.\footnote{We use a RoBERTa classifier~\cite{liu2019roberta} trained on MultiNLI~\cite{williams2018broad} as entailment scorer.}

\begin{table*}[t]
\begin{center}
\begin{small}
\begin{tabular}{lcccccc}
\toprule
Models & BLEU-4 & METEOR & ROUGE-L & Q-BLEU1 & QPP & QAP  \\
\midrule
\multicolumn{7}{c}{\textsc{Baselines}} \\
\midrule
Cross-Entropy~\cite{zhang2019addressing}  & 17.88 & 22.38 & 46.39 & 49.01 & 28.83  & 54.25 \\
ROUGE-RL  & 18.03 & 22.55 & 46.64 & 49.52 & 29.09 & 55.07 \\
QPP-RL  & 17.90 & 22.55 & 46.68 & 49.50 & 30.10 & 55.50  \\
QAP-RL  & 18.22 & 22.69 & 46.65 & 49.72 & 30.03  & \textbf{57.60}  \\
\midrule
\multicolumn{7}{c}{\textsc{Multi-Reward Models}} \\
\midrule
\newcite{pasunuru2018multi}$^\dagger$ & 18.36 & 22.55 & 46.75 & 49.66 & 30.03 & 56.51  \\
\hdashline
Our SM-Bandit$^\dagger$ & \textbf{18.68} & \textbf{22.88} & 46.80 & \textbf{50.02} & \textbf{30.15}  & 56.92 \\
Our HM-Bandit$^\dagger$ & 18.55 & 22.82 & \textbf{46.84} & 50.01 & 30.07 & 56.78 \\
\bottomrule
\end{tabular}
\end{small}
\end{center}
\vspace{-9pt}
\caption{Performance of our baselines and multi-armed bandit-based models on question generation task. $\dagger$ denotes that these models use ROUGE-L, QPP, and QAP rewards during the optimization.
\label{table:question-generation-results}
}
\vspace{-8pt}
\end{table*}

\paragraph{Dataset \& Evaluation.}
We use the WebNLG dataset~\cite{gardent2017webnlg} - a widely used English benchmark for data-to-text generation which focuses on micro-planning involving several subtasks like referring expression generation, aggregation, lexicalization, sentence segmentation, and surface realization. It contains 9,674 unique RDF triple-sets and 25,298 text references, which is divided into train, dev, and test sets.\footnote{\url{https://webnlg-challenge.loria.fr/}} We report all our results on the `seen' and `unseen' part of the test set. For each sample, the input is a set of up to 7 RDF triples from DBPedia, and the output is their text descriptions. 
The standard evaluation metrics for this dataset include METEOR\footnote{\url{http://www.cs.cmu.edu/~alavie/METEOR/}}~\cite{denkowski2014meteor}, BLEU~\cite{papineni2002bleu}, and TER\footnote{\url{http://www.cs.umd.edu/~snover/tercom/}}~\cite{snover2006study}.
We also report ROUGE-L~\cite{lin2004rouge} and Entailment-Score~\cite{pasunuru2018multi}.

\subsection{Training Details}
All the hyperparameters are tuned on the validation set for both question generation and data-to-text tasks. We use TITAN X and GeForce GTX 1080 GPUs for all our experiments.
For the question generation task, we use two layers for both encoder and decoder. We set the hidden size of LSTM-RNN to 600 and use BERT-based contextual embeddings as input. We use a batch size of 32, encoder maximum length of 512 and decoder maximum length of 50, and maximum gradient clipping of~5. We use Adam optimizer~\cite{kingma2014adam} with a learning rate of 1e-3 and 1e-6 for the cross-entropy and RL models, respectively.
For data-to-text task, we use the same hyperparameters as discussed in~\citet{zhao2020bridging} for the cross-entropy model, e.g., we use Adam with a batch size of 64 and an initial learning rate of 0.001. All RL models are initialized with the best cross-entropy model checkpoint, and use Adam with a learning rate of 1e-6.
We refer to Appendix~\ref{sec:supp-training-details} for full training details.

\begin{table*}[t]
\begin{center}
\begin{small}
\begin{tabular}{lccccc}
\toprule
Models & BLEU ($\uparrow$) & METEOR ($\uparrow$) & TER ($\downarrow$) & ROUGE-L ($\uparrow$) & Entailment ($\uparrow$) \\
\midrule
\multicolumn{6}{c}{\textsc{Baselines}} \\
\midrule
Cross-Entropy~\cite{zhao2020bridging} & 63.14/22.56 & 44.85/27.79 & 33.97/71.02 & 74.25/49.83 & 99.27/88.16 \\
ROUGE-RL  & 63.35/22.96 & 44.84/28.18 & 33.85/70.58 & 74.29/50.07 & 99.11/88.59  \\
BLEU-RL  & 63.24/23.06 & 44.82/28.21 & 33.94/70.33 & 74.26/50.00 & 99.30/87.82 \\
Ent-RL  & 63.28/22.49 & 44.96/28.11 & 34.03/72.29 & 74.29/49.99 & 99.84/90.57 \\
\midrule
\multicolumn{6}{c}{\textsc{Multi-Reward Models}} \\
\midrule
\newcite{pasunuru2018multi}$^\dagger$ & 63.00/22.58 & 45.03/28.29 & 34.22/72.71 & 74.29/50.05 & 99.56/91.83  \\
\hdashline
Our SM-Bandit$^\dagger$  & \textbf{63.46}/\textbf{23.23} & \textbf{45.37}/28.68 & 33.59/\textbf{70.21} & 74.38/\textbf{50.30} & 100.13/92.35 \\
Our HM-Bandit$^\dagger$  & 63.38/23.21 & 45.34/\textbf{28.70} & \textbf{33.58}/70.17 & \textbf{74.39}/50.26 & \textbf{100.21}/\textbf{92.40} \\
\bottomrule
\end{tabular}
\end{small}
\end{center}
\vspace{-10pt}
\caption{Performance of our baselines and multi-arm bandit-based models on the `seen/unseen' test set of WebNLG data-to-text task. The unseen set has categories that are not seen during the training, hence can be consider as a test-only transfer setup. $\dagger$ denotes that these models use ROUGE-L, BLEU, and Entailment rewards during the optimization. For TER metric, lower ($\downarrow$) is better. For all other metrics, higher ($\uparrow$) is better.
\label{table:e2e-results}
\vspace{-10pt}
}
\end{table*}

%%%%%%%%%%%%%%%%%%%%%%%%Results Section%%%%%%%%%%%%%%%%%%%%%%%%%%%%%%%%%%%%%%%%
%%%%%%%%%%%%%%%%%%%%%%%%%%%%%%%%%%%%%%%%%%%%%%%%%%%%%%%%%%%%%%%%%%%%

\section{Results and Analysis}
In this section, we present the performance of previous work, our cross-entropy baselines, our RL-based baselines, and finally our multi-arm bandit-based models. We start with results on automatic evaluation (Sec.~\ref{subsec:results-qg}-\ref{subsec:results-e2e}). Next, we present results on human evaluation (Sec.~\ref{subsec-human-evaluation}). Finally, we present an interpretable analysis on the bandits (Sec.~\ref{subsec:bandit-analysis}).

\subsection{Results on Question Generation}
\label{subsec:results-qg}
\paragraph{Baselines.}
Table~\ref{table:question-generation-results} presents results on the question generation dataset for our baselines. We use the previous state-of-the-art work~\cite{zhang2019addressing} as our cross-entropy baseline. Next, we apply policy gradients-based reinforcement learning (RL) approach, and observe that all these models are better than the baseline in all metrics. Next, we will discuss the multi-reward RL models.

\paragraph{Multi-Armed Bandit Approaches.}
Finally, we evaluate our two bandit approaches: SM-Bandit and HM-Bandit as described in Sec.~\ref{subsec:sm-bandit} and Sec.~\ref{subsec:hm-bandit}, respectively. Further, for a fair comparison of our multi-arm bandit-based models, we further implemented multi-reward alternate optimization approach introduced by~\newcite{pasunuru2018multi} and considered it as baseline for our multi-reward models.\footnote{We do not compare with fixed weighted combination of metrics during RL optimization, as finding the optimal weighted combination is exponential complexity (searching among $100$ values for $n$ metrics needs $100^n$ tuning experiments), which we want to avoid via our bandit approach.}\textsuperscript{,}\footnote{We also experimented with the random choice of metrics during optimization. The results on the question generation task are very close to the baseline model~\cite{pasunuru2018multi}: 18.31(BLEU), 22.50 (METEOR), 46.75 (ROUGE-L), 49.65 (Q-BLEU1), 30.04 (QPP), 56.56 (QAP). This is expected as the random choice baseline is same as uniform sampling of metrics, which is closer to alternate optimization.}
This model is slightly better than single reward-based RL baselines.  Table~\ref{table:question-generation-results} presents the performance of the proposed two bandit models (SM-Bandit and HM-Bandit) on various automatic evaluation metrics, and we observe that on average these models perform much better than the cross-entropy and single reward RL baseline models. Further, our bandit models also perform better than the multi-reward approach proposed by~\newcite{pasunuru2018multi}, suggesting that our bandit-based models are able to dynamically select the reward to optimize for overall improvement in all the metrics that we want to optimize. Also see discussion of significant improvements in human evaluation in Sec~\ref{subsec-human-evaluation}.

\begin{figure*}[t]
\centering
\includegraphics[width=0.9\linewidth]{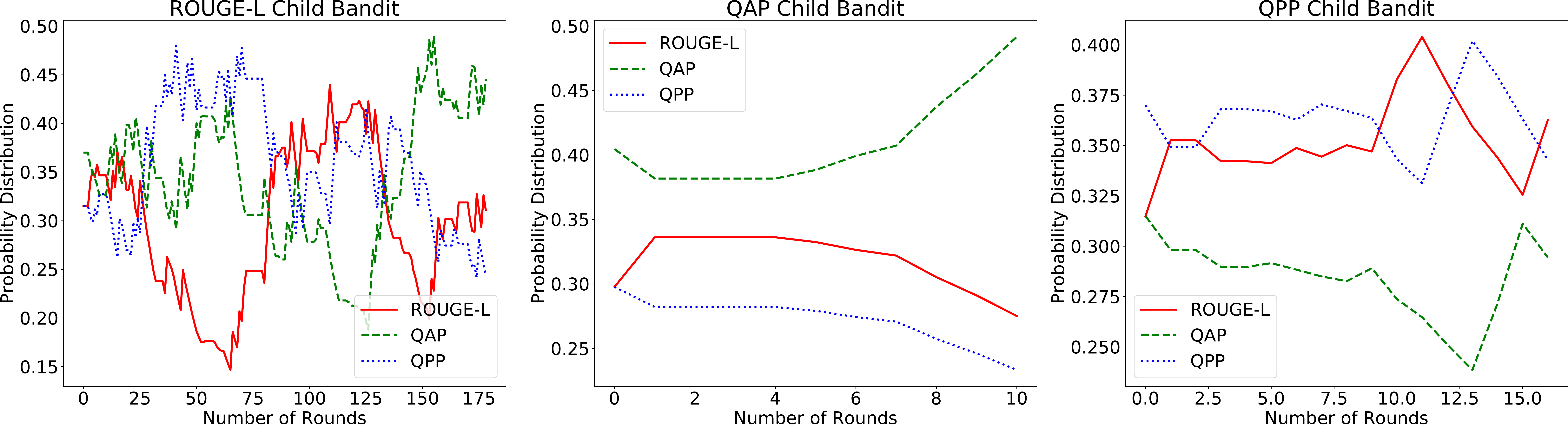}
\vspace{-7pt}
\caption{Plots showing the probability distribution of each child bandit of the HM-Bandit model on the QG task.
\label{fig:qg-hm-bandit-plot}
\vspace{-10pt}
}
\end{figure*}

\subsection{Results on Data-to-Text Generation}
\label{subsec:results-e2e}
\paragraph{Baselines.}
Table~\ref{table:e2e-results} presents our baselines on the WebNLG data-to-text task. Our cross-entropy model is comparable to the very recent state-of-the-art model~\cite{zhao2020bridging}. Further, we present single reward based RL models with ROUGE-L, BLEU, and Entailment score as rewards, which again perform better than our cross-entropy model. Next, we will discuss multi-reward models.

\paragraph{Multi-Armed Bandit Approaches.}
Table~\ref{table:e2e-results} also presents our multi-armed bandit models (SM-Bandit and HM-Bandit) which simultaneously use ROUGE-L, BLEU, and Entailment score as rewards. Again, we consider the model proposed by~\newcite{pasunuru2018multi} as a baseline for multi-reward models. On average, our bandit-based models perform better than all our baselines that are discussed in the above paragraph and also the model based on~\newcite{pasunuru2018multi}.\footnote{In general, we observe better improvements with our bandit-based models in the unseen-test transfer setup.} Also see discussion of significant improvements in human evaluation in Sec~\ref{subsec-human-evaluation}.

\begin{table}[t]
\begin{center}
\begin{small}
\begin{tabular}{lcccc}
\toprule
Model & ROUGE & PB~\shortcite{pasunuru2018multi} & SMB & HMB\\
\midrule
\multicolumn{5}{c}{\textsc{Question Generation Task}} \\
\midrule
Relevance & 4.28 & 4.40 & 4.56 & 4.55 \\
Coherence & 4.42 & 4.48 & 4.49 & 4.47 \\
\midrule
\multicolumn{5}{c}{\textsc{WebNLG Data-to-Text Task}} \\
\midrule
Relevance & 4.61 & 4.68 & 4.79 & 4.81 \\
Coherence & 4.75 & 4.79 & 4.78 & 4.80 \\
\bottomrule
\end{tabular}
\end{small}
\end{center}
\vspace{-10pt}
\caption{Human evaluation results on QG and WebNLG tasks. ROUGE: ROUGE-L single-reward RL; PB~\shortcite{pasunuru2018multi}:~\citet{pasunuru2018multi}. Our SM-Bandit and HM-Bandit are statistically significantly better than ROUGE and PB models (see Sec.~\ref{subsec-human-evaluation}).
\label{table:human-evaluation-results}
\vspace{-10pt}
}
\end{table}

\subsection{Human Evaluation}
\label{subsec-human-evaluation}
It is shown that RL models can game the metric that we use as the objective function~\cite{paulus2017deep}. This motivated us to optimize the RL models on multiple metrics simultaneously, thus trying to improve all the metrics and making the RL model hard to game any particular metric. In this section, we validate the superiority of our bandit models via human evaluation studies. 

We performed anonymous human evaluation studies using Amazon Mechanical Turk (MTurk). We chose human annotators such that they are located in the USA, have at least 10,000 approved HITs, and have an approval rate of greater than 98\%. For both question generation and WebNLG data-to-text, we considered 200 samples for each, and compared ROUGE-L RL,~\newcite{pasunuru2018multi}, SM-Bandit, and HM-Bandit models by asking the annotators to rate the quality of the generated outputs based on relevance and coherence on 5-point Likert scale.\footnote{For question generation, relevance is defined as how clearly the generated question will be able to point to the right answer, given an input paragraph as context. For WebNLG data-to-text, relevance is defined as how related is the generated description w.r.t. the given RDF data such as mentioning the facts. For both tasks, coherence is based on the logic, readability, and fluency of the generated question or description.}
Table~\ref{table:human-evaluation-results} presents these human evaluation studies. In terms of relevance, our SM-Bandit and HM-Bandit models are significantly better than~\newcite{pasunuru2018multi} ($p{<}0.01$) and ROUGE-L RL models ($p{<}0.01$) on question generation, while maintaining coherence.\footnote{We use bootstrap test~\cite{efron1994introduction,noreen1989computer} for calculating the statistical significance score.}
On data-to-text, in terms of relevance, our SM-Bandit and HM-Bandit models are significantly better than~\citet{pasunuru2017multi} with $p{<}0.03$  and $p{<}0.02$, respectively. Also, both bandit models are significantly better than ROUGE-L RL model with $p{<}0.01$.
We also performed a similar human evaluation study for the test-only transfer setup on the unseen WebNLG test set, and the results are in Table~\ref{table:human-evaluation-transfer-results}. Here also our bandit-based model (HM-Bandit) performed statistically significantly better than~\newcite{pasunuru2018multi} on relevance metric with $p<0.01$, while maintaining coherence.

\begin{table}[t]
\begin{center}
\begin{small}
\begin{tabular}{lcc}
\toprule
Model & \newcite{pasunuru2018multi} & HM-Bandit\\
\midrule
Relevance & 3.49 & 3.68  \\
Coherence & 3.44 & 3.46  \\
\bottomrule
\end{tabular}
\end{small}
\end{center}
\vspace{-7pt}
\caption{Human evaluation results on WebNLG `unseen' test set. Our HM-Bandit is statistically significantly better than~\newcite{pasunuru2018multi} on relevance metric.
\label{table:human-evaluation-transfer-results}
\vspace{-10pt}
}
\end{table}

\subsection{Interpretable Bandit Analysis}
\label{subsec:bandit-analysis}
\begin{figure}[t]
\centering
\includegraphics[width=0.85\linewidth]{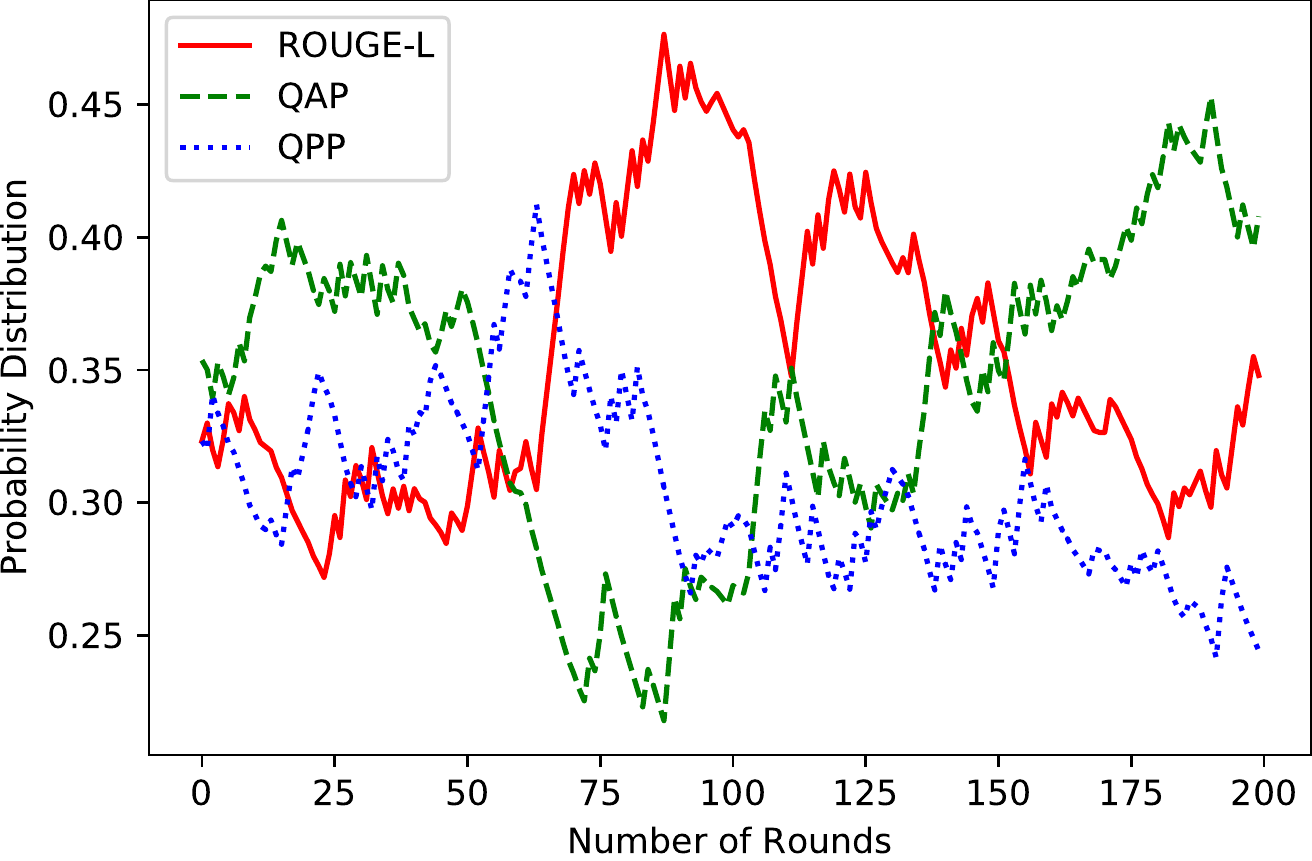}
\vspace{-5pt}
\caption{Plot showing the probability distribution of each arm of the SM-Bandit on question generation task. 
\label{fig:qg-sm-bandit-plot}
}
\vspace{-15pt}
\end{figure}

Figure~\ref{fig:qg-sm-bandit-plot} presents the interpretable visualization of the probability distribution of each arm of the SM-Bandit as the training progresses. We observe that each metric has played an important role (as high probability arm) for at least a few rounds over the training trajectory. Also, there are multiple switchings of these metrics over the training trajectory, suggesting that this kind of automatic dynamic switching is important to improve the overall performance of RL models with multiple rewards. 

Figure~\ref{fig:qg-hm-bandit-plot} presents the progress of child bandits of HM-Bandit during the training for question generation. As discussed in Sec.~\ref{subsec:hm-bandit}, these child bandits are controlled by a controller that selects the under-performing bandit. We observe that our HM-Bandit mostly used ROUGE-L child bandit for overall improvement in all metrics (as it is the under-performing metric). Further, each child bandit gave more importance to the metric that it wants to improve, e.g., the QAP child bandit gave more importance to the QAP arm. However, there is an exception for the ROUGE-L child bandit, where ROUGE-L arm is not the most important, suggesting that to improve the ROUGE-L metric other RL loss functions (QAP and QPP) are also useful.

%%%%%%%%%%%%%%%%%%%%%%%%Results Section%%%%%%%%%%%%%%%%%%%%%%%%%%%%%%%%%%%%%%%%
%%%%%%%%%%%%%%%%%%%%%%%%%%%%%%%%%%%%%%%%%%%%%%%%%%%%%%%%%%%%%%%%%%%%
% \input{analysis.tex}

%%%%%%%%%%%%%%%%%%%%%%%%Conclusion%%%%%%%%%%%%%%%%%%%%%%%%%%%%%%%%%%%%%%%%
%%%%%%%%%%%%%%%%%%%%%%%%%%%%%%%%%%%%%%%%%%%%%%%%%%%%%%%%%%%%%%%%%%%%
\section{Conclusion}
We presented novel approaches for dynamically optimizing multiple reward metrics simultaneously via multi-armed bandit approach in the context of language generation. We described two such mechanisms, namely single bandit and hierarchical bandit with multiple rewards. We conducted experiments on two challenging language generation tasks: question generation and data-to-text generation, and our method achieved strong improvements based on human evaluation over previous approaches. We further presented interpretable analysis on our bandit methods.

\section*{Acknowledgments}
We thank the reviewers for their helpful comments, Shiyue Zhang for the code to build QPP and QAP rewards, and Chao Zhao for the code of their SOTA WebNLG model. This work was supported by DARPA YFA17-D17AP00022, NSF-CAREER Award 1846185, ONR
Grant N00014-18-1-2871, and Microsoft PhD Fellowship. The views contained in this article are those of the authors and not of the funding agency.

\bibliography{citations}

\begin{thebibliography}{106}
\expandafter\ifx\csname natexlab\endcsname\relax\def\natexlab#1{#1}\fi

\bibitem[{Audibert et~al.(2009)Audibert, Munos, and
  Szepesv{\'a}ri}]{audibert2009exploration}
Jean-Yves Audibert, R{\'e}mi Munos, and Csaba Szepesv{\'a}ri. 2009.
\newblock Exploration--exploitation tradeoff using variance estimates in
  multi-armed bandits.
\newblock \emph{Theoretical Computer Science}, 410(19):1876--1902.

\bibitem[{Auer et~al.(2002{\natexlab{a}})Auer, Cesa-Bianchi, and
  Fischer}]{auer2002finite}
Peter Auer, Nicolo Cesa-Bianchi, and Paul Fischer. 2002{\natexlab{a}}.
\newblock Finite-time analysis of the multiarmed bandit problem.
\newblock \emph{Machine learning}, 47(2-3):235--256.

\bibitem[{Auer et~al.(2002{\natexlab{b}})Auer, Cesa-Bianchi, Freund, and
  Schapire}]{auer2002nonstochastic}
Peter Auer, Nicolo Cesa-Bianchi, Yoav Freund, and Robert~E Schapire.
  2002{\natexlab{b}}.
\newblock The nonstochastic multiarmed bandit problem.
\newblock \emph{SIAM journal on computing}, 32(1):48--77.

\bibitem[{Bahdanau et~al.(2015)Bahdanau, Cho, and Bengio}]{bahdanau2014neural}
Dzmitry Bahdanau, Kyunghyun Cho, and Yoshua Bengio. 2015.
\newblock Neural machine translation by jointly learning to align and
  translate.
\newblock In \emph{ICLR}.

\bibitem[{Bubeck et~al.(2012)Bubeck, Cesa-Bianchi et~al.}]{bubeck2012regret}
S{\'e}bastien Bubeck, Nicolo Cesa-Bianchi, et~al. 2012.
\newblock Regret analysis of stochastic and nonstochastic multi-armed bandit
  problems.
\newblock \emph{Foundations and Trends{\textregistered} in Machine Learning},
  5(1):1--122.

\bibitem[{Burtini et~al.(2015)Burtini, Loeppky, and
  Lawrence}]{burtini2015survey}
Giuseppe Burtini, Jason Loeppky, and Ramon Lawrence. 2015.
\newblock A survey of online experiment design with the stochastic multi-armed
  bandit.
\newblock \emph{arXiv preprint arXiv:1510.00757}.

\bibitem[{Celikyilmaz et~al.(2018)Celikyilmaz, Bosselut, He, and
  Choi}]{celikyilmaz2018deep}
Asli Celikyilmaz, Antoine Bosselut, Xiaodong He, and Yejin Choi. 2018.
\newblock Deep communicating agents for abstractive summarization.
\newblock In \emph{NAACL}, pages 1662--1675.

\bibitem[{Chapelle and Li(2011)}]{chapelle2011empirical}
Olivier Chapelle and Lihong Li. 2011.
\newblock An empirical evaluation of thompson sampling.
\newblock In \emph{Advances in neural information processing systems}, pages
  2249--2257.

\bibitem[{Chen and Mooney(2008)}]{chen2008learning}
David~L Chen and Raymond~J Mooney. 2008.
\newblock Learning to sportscast: a test of grounded language acquisition.
\newblock In \emph{Proceedings of the 25th international conference on Machine
  learning}, pages 128--135. ACM.

\bibitem[{Chen et~al.(2013)Chen, Wang, and Yuan}]{chen2013combinatorial}
Wei Chen, Yajun Wang, and Yang Yuan. 2013.
\newblock Combinatorial multi-armed bandit: General framework and applications.
\newblock In \emph{International Conference on Machine Learning}, pages
  151--159.

\bibitem[{Chen and Bansal(2018)}]{chen2018fast}
Yen-Chun Chen and Mohit Bansal. 2018.
\newblock Fast abstractive summarization with reinforce-selected sentence
  rewriting.
\newblock In \emph{Proceedings of the 56th Annual Meeting of the Association
  for Computational Linguistics (Volume 1: Long Papers)}, volume~1, pages
  675--686.

\bibitem[{Chopra et~al.(2016)Chopra, Auli, and Rush}]{chopra2016abstractive}
Sumit Chopra, Michael Auli, and Alexander~M Rush. 2016.
\newblock Abstractive sentence summarization with attentive recurrent neural
  networks.
\newblock In \emph{Proceedings of the 2016 Conference of the North American
  Chapter of the Association for Computational Linguistics: Human Language
  Technologies}, pages 93--98.

\bibitem[{Denkowski and Lavie(2014)}]{denkowski2014meteor}
Michael Denkowski and Alon Lavie. 2014.
\newblock Meteor universal: Language specific translation evaluation for any
  target language.
\newblock In \emph{Proceedings of the ninth workshop on statistical machine
  translation}, pages 376--380.

\bibitem[{Devlin et~al.(2019)Devlin, Chang, Lee, and
  Toutanova}]{devlin2019bert}
Jacob Devlin, Ming-Wei Chang, Kenton Lee, and Kristina Toutanova. 2019.
\newblock {BERT}: Pre-training of deep bidirectional transformers for language
  understanding.
\newblock In \emph{Proceedings of the 2019 Conference of the North American
  Chapter of the Association for Computational Linguistics: Human Language
  Technologies, Volume 1 (Long and Short Papers)}, pages 4171--4186.

\bibitem[{Distiawan et~al.(2018)Distiawan, Qi, Zhang, and
  Wang}]{distiawan2018gtr}
Bayu Distiawan, Jianzhong Qi, Rui Zhang, and Wei Wang. 2018.
\newblock Gtr-lstm: A triple encoder for sentence generation from rdf data.
\newblock In \emph{Proceedings of the 56th Annual Meeting of the Association
  for Computational Linguistics (Volume 1: Long Papers)}, pages 1627--1637.

\bibitem[{Du and Cardie(2018)}]{du2018harvesting}
Xinya Du and Claire Cardie. 2018.
\newblock Harvesting paragraph-level question-answer pairs from wikipedia.
\newblock In \emph{Proceedings of the 56th Annual Meeting of the Association
  for Computational Linguistics (Volume 1: Long Papers)}, pages 1907--1917.

\bibitem[{Du et~al.(2017)Du, Shao, and Cardie}]{du2017learning}
Xinya Du, Junru Shao, and Claire Cardie. 2017.
\newblock Learning to ask: Neural question generation for reading
  comprehension.
\newblock In \emph{Proceedings of the 55th Annual Meeting of the Association
  for Computational Linguistics (Volume 1: Long Papers)}, volume~1, pages
  1342--1352.

\bibitem[{Duh et~al.(2012)Duh, Sudoh, Wu, Tsukada, and
  Nagata}]{duh2012learning}
Kevin Duh, Katsuhito Sudoh, Xianchao Wu, Hajime Tsukada, and Masaaki Nagata.
  2012.
\newblock Learning to translate with multiple objectives.
\newblock In \emph{Proceedings of the 50th Annual Meeting of the Association
  for Computational Linguistics (Volume 1: Long Papers)}, pages 1--10.

\bibitem[{Du{\v{s}}ek and Jurcicek(2016)}]{duvsek2016sequence}
Ond{\v{r}}ej Du{\v{s}}ek and Filip Jurcicek. 2016.
\newblock Sequence-to-sequence generation for spoken dialogue via deep syntax
  trees and strings.
\newblock In \emph{Proceedings of the 54th Annual Meeting of the Association
  for Computational Linguistics (Volume 2: Short Papers)}, volume~2, pages
  45--51.

\bibitem[{Du{\v{s}}ek et~al.(2020)Du{\v{s}}ek, Novikova, and
  Rieser}]{dusek2019e2e}
Ond{\v{r}}ej Du{\v{s}}ek, Jekaterina Novikova, and Verena Rieser. 2020.
\newblock Evaluating the state-of-the-art of end-to-end natural language
  generation: {The} {E2E} {NLG} {Challenge}.
\newblock \emph{Computer Speech \& Language}, 59:123--156.

\bibitem[{Efron and Tibshirani(1994)}]{efron1994introduction}
Bradley Efron and Robert~J Tibshirani. 1994.
\newblock \emph{An introduction to the bootstrap}.
\newblock CRC press.

\bibitem[{Gardent et~al.(2017)Gardent, Shimorina, Narayan, and
  Perez-Beltrachini}]{gardent2017webnlg}
Claire Gardent, Anastasia Shimorina, Shashi Narayan, and Laura
  Perez-Beltrachini. 2017.
\newblock The {WebNLG} challenge: Generating text from rdf data.
\newblock In \emph{Proceedings of the 10th International Conference on Natural
  Language Generation}, pages 124--133.

\bibitem[{Gehrmann et~al.(2018)Gehrmann, Dai, Elder, and
  Rush}]{gehrmann2018end}
Sebastian Gehrmann, Falcon Dai, Henry Elder, and Alexander~M Rush. 2018.
\newblock End-to-end content and plan selection for data-to-text generation.
\newblock In \emph{Proceedings of the 11th International Conference on Natural
  Language Generation}, pages 46--56.

\bibitem[{Gong(2018)}]{gong2018technical}
Heng Gong. 2018.
\newblock Technical report for e2e nlg challenge.
\newblock \emph{E2E NLG Challenge System Descriptions}.

\bibitem[{Graves et~al.(2017)Graves, Bellemare, Menick, Munos, and
  Kavukcuoglu}]{graves2017automated}
Alex Graves, Marc~G Bellemare, Jacob Menick, Remi Munos, and Koray Kavukcuoglu.
  2017.
\newblock Automated curriculum learning for neural networks.
\newblock In \emph{Proceedings of the 34th International Conference on Machine
  Learning-Volume 70}, pages 1311--1320. JMLR. org.

\bibitem[{Gu et~al.(2017)Gu, Cho, and Li}]{gu2017trainable}
Jiatao Gu, Kyunghyun Cho, and Victor~OK Li. 2017.
\newblock Trainable greedy decoding for neural machine translation.
\newblock In \emph{Proceedings of the 2017 Conference on Empirical Methods in
  Natural Language Processing}, pages 1968--1978.

\bibitem[{Gu et~al.(2016)Gu, Lu, Li, and Li}]{gu2016incorporating}
Jiatao Gu, Zhengdong Lu, Hang Li, and Victor~OK Li. 2016.
\newblock Incorporating copying mechanism in sequence-to-sequence learning.
\newblock In \emph{Proceedings of the 54th Annual Meeting of the Association
  for Computational Linguistics (Volume 1: Long Papers)}, pages 1631--1640.

\bibitem[{Gulcehre et~al.(2016)Gulcehre, Ahn, Nallapati, Zhou, and
  Bengio}]{gulcehre2016pointing}
Caglar Gulcehre, Sungjin Ahn, Ramesh Nallapati, Bowen Zhou, and Yoshua Bengio.
  2016.
\newblock Pointing the unknown words.
\newblock In \emph{Proceedings of the 54th Annual Meeting of the Association
  for Computational Linguistics (Volume 1: Long Papers)}, volume~1, pages
  140--149.

\bibitem[{Guo et~al.(2018)Guo, Pasunuru, and Bansal}]{guo2018dynamic}
Han Guo, Ramakanth Pasunuru, and Mohit Bansal. 2018.
\newblock Dynamic multi-level multi-task learning for sentence simplification.
\newblock In \emph{Proceedings of the 27th International Conference on
  Computational Linguistics}, pages 462--476.

\bibitem[{Guo et~al.(2019{\natexlab{a}})Guo, Pasunuru, and
  Bansal}]{guo2019autosem}
Han Guo, Ramakanth Pasunuru, and Mohit Bansal. 2019{\natexlab{a}}.
\newblock {AutoSeM}: Automatic task selection and mixing in multi-task
  learning.
\newblock In \emph{Proceedings of the 2019 Conference of the North American
  Chapter of the Association for Computational Linguistics: Human Language
  Technologies, Volume 1 (Long and Short Papers)}, pages 3520--3531.

\bibitem[{Guo et~al.(2019{\natexlab{b}})Guo, Zhang, Teng, and
  Lu}]{guo2019densely}
Zhijiang Guo, Yan Zhang, Zhiyang Teng, and Wei Lu. 2019{\natexlab{b}}.
\newblock Densely connected graph convolutional networks for graph-to-sequence
  learning.
\newblock \emph{Transactions of the Association for Computational Linguistics},
  7:297--312.

\bibitem[{Hochreiter and Schmidhuber(1997)}]{hochreiter1997long}
Sepp Hochreiter and J{\"u}rgen Schmidhuber. 1997.
\newblock Long short-term memory.
\newblock \emph{Neural computation}, 9(8):1735--1780.

\bibitem[{Kaelbling et~al.(1996)Kaelbling, Littman, and
  Moore}]{kaelbling1996reinforcement}
Leslie~Pack Kaelbling, Michael~L Littman, and Andrew~W Moore. 1996.
\newblock Reinforcement learning: A survey.
\newblock \emph{Journal of artificial intelligence research}, 4:237--285.

\bibitem[{Kim et~al.(2019)Kim, Lee, Shin, and Jung}]{kim2018improving}
Yanghoon Kim, Hwanhee Lee, Joongbo Shin, and Kyomin Jung. 2019.
\newblock Improving neural question generation using answer separation.
\newblock In \emph{Proceedings of the AAAI Conference on Artificial
  Intelligence}, volume~33, pages 6602--6609.

\bibitem[{Kingma and Ba(2015)}]{kingma2014adam}
Diederik Kingma and Jimmy Ba. 2015.
\newblock Adam: A method for stochastic optimization.
\newblock In \emph{ICLR}.

\bibitem[{Kumar et~al.(2019)Kumar, Ramakrishnan, and Li}]{kumar2018framework}
Vishwajeet Kumar, Ganesh Ramakrishnan, and Yuan-Fang Li. 2019.
\newblock Putting the horse before the cart: A generator-evaluator framework
  for question generation from text.
\newblock In \emph{Proceedings of the 23rd Conference on Computational Natural
  Language Learning (CoNLL)}, pages 812--821.

\bibitem[{Kveton et~al.(2019)Kveton, Szepesvari, Vaswani, Wen, Lattimore, and
  Ghavamzadeh}]{kveton2018garbage}
Branislav Kveton, Csaba Szepesvari, Sharan Vaswani, Zheng Wen, Tor Lattimore,
  and Mohammad Ghavamzadeh. 2019.
\newblock Garbage in, reward out: Bootstrapping exploration in multi-armed
  bandits.
\newblock In \emph{International Conference on Machine Learning}, pages
  3601--3610.

\bibitem[{Lampouras and Vlachos(2016)}]{lampouras2016imitation}
Gerasimos Lampouras and Andreas Vlachos. 2016.
\newblock Imitation learning for language generation from unaligned data.
\newblock In \emph{Proceedings of COLING 2016, the 26th International
  Conference on Computational Linguistics: Technical Papers}, pages 1101--1112.

\bibitem[{Lattimore and Szepesv\'{a}ri(2019)}]{LS19bandit-book}
Tor Lattimore and Csaba Szepesv\'{a}ri. 2019.
\newblock \emph{Bandit Algorithms}.
\newblock Cambridge University Press (preprint).

\bibitem[{Lebret et~al.(2016)Lebret, Grangier, and Auli}]{lebret2016neural}
R{\'e}mi Lebret, David Grangier, and Michael Auli. 2016.
\newblock Neural text generation from structured data with application to the
  biography domain.
\newblock In \emph{Proceedings of the 2016 Conference on Empirical Methods in
  Natural Language Processing}, pages 1203--1213.

\bibitem[{Li et~al.(2016)Li, Monroe, Ritter, Jurafsky, Galley, and
  Gao}]{li2016deep}
Jiwei Li, Will Monroe, Alan Ritter, Dan Jurafsky, Michel Galley, and Jianfeng
  Gao. 2016.
\newblock Deep reinforcement learning for dialogue generation.
\newblock In \emph{Proceedings of the 2016 Conference on Empirical Methods in
  Natural Language Processing}, pages 1192--1202.

\bibitem[{Li et~al.(2010)Li, Chu, Langford, and Schapire}]{li2010contextual}
Lihong Li, Wei Chu, John Langford, and Robert~E Schapire. 2010.
\newblock A contextual-bandit approach to personalized news article
  recommendation.
\newblock In \emph{Proceedings of the 19th international conference on World
  wide web}, pages 661--670. ACM.

\bibitem[{Li et~al.(2018)Li, Jamieson, DeSalvo, Rostamizadeh, and
  Talwalkar}]{li2016hyperband}
Lisha Li, Kevin Jamieson, Giulia DeSalvo, Afshin Rostamizadeh, and Ameet
  Talwalkar. 2018.
\newblock Hyperband: A novel bandit-based approach to hyperparameter
  optimization.
\newblock \emph{Journal of Machine Learning Research}, 18(185):1--52.

\bibitem[{Liang et~al.(2009)Liang, Jordan, and Klein}]{liang2009learning}
Percy Liang, Michael~I Jordan, and Dan Klein. 2009.
\newblock Learning semantic correspondences with less supervision.
\newblock In \emph{Proceedings of the Joint Conference of the 47th Annual
  Meeting of the ACL and the 4th International Joint Conference on Natural
  Language Processing of the AFNLP: Volume 1-Volume 1}, pages 91--99.
  Association for Computational Linguistics.

\bibitem[{Lin(2004)}]{lin2004rouge}
Chin-Yew Lin. 2004.
\newblock {ROUGE}: A package for automatic evaluation of summaries.
\newblock In \emph{Proceedings of Workshop on Text Summarization Branches Out,
  Conference Workshop of ACL}.

\bibitem[{Liu et~al.(2019{\natexlab{a}})Liu, Zhao, Niu, Lai, He, Wei, and
  Xu}]{liu2019learning}
Bang Liu, Mingjun Zhao, Di~Niu, Kunfeng Lai, Yancheng He, Haojie Wei, and
  Yu~Xu. 2019{\natexlab{a}}.
\newblock Learning to generate questions by learningwhat not to generate.
\newblock In \emph{The World Wide Web Conference}, pages 1106--1118.

\bibitem[{Liu et~al.(2017)Liu, Zhu, Ye, Guadarrama, and
  Murphy}]{liu2017improved}
Siqi Liu, Zhenhai Zhu, Ning Ye, Sergio Guadarrama, and Kevin Murphy. 2017.
\newblock Improved image captioning via policy gradient optimization of spider.
\newblock In \emph{Proceedings of the IEEE international conference on computer
  vision}, pages 873--881.

\bibitem[{Liu et~al.(2019{\natexlab{b}})Liu, Ott, Goyal, Du, Joshi, Chen, Levy,
  Lewis, Zettlemoyer, and Stoyanov}]{liu2019roberta}
Yinhan Liu, Myle Ott, Naman Goyal, Jingfei Du, Mandar Joshi, Danqi Chen, Omer
  Levy, Mike Lewis, Luke Zettlemoyer, and Veselin Stoyanov. 2019{\natexlab{b}}.
\newblock {RoBERTa}: A robustly optimized {BERT} pretraining approach.
\newblock \emph{arXiv preprint arXiv:1907.11692}.

\bibitem[{Luong et~al.(2015)Luong, Pham, and Manning}]{luong2015effective}
Thang Luong, Hieu Pham, and Christopher~D Manning. 2015.
\newblock Effective approaches to attention-based neural machine translation.
\newblock In \emph{Proceedings of the 2015 Conference on Empirical Methods in
  Natural Language Processing}, pages 1412--1421.

\bibitem[{Macready and Wolpert(1998)}]{macready1998bandit}
William~G Macready and David~H Wolpert. 1998.
\newblock Bandit problems and the exploration/exploitation tradeoff.
\newblock \emph{IEEE Transactions on evolutionary computation}, 2(1):2--22.

\bibitem[{Marcheggiani and Perez-Beltrachini(2018)}]{marcheggiani2018deep}
Diego Marcheggiani and Laura Perez-Beltrachini. 2018.
\newblock Deep graph convolutional encoders for structured data to text
  generation.
\newblock In \emph{INLG}.

\bibitem[{Mei et~al.(2016)Mei, UChicago, Bansal, and Walter}]{mei2015talk}
Hongyuan Mei, TTI UChicago, Mohit Bansal, and Matthew~R Walter. 2016.
\newblock What to talk about and how? selective generation using lstms with
  coarse-to-fine alignment.
\newblock In \emph{Proceedings of NAACL-HLT}, pages 720--730.

\bibitem[{Merentitis et~al.(2018)Merentitis, Rasul, Vollgraf, Sheikh, and
  Bergmann}]{merentitis2019bandit}
Andreas Merentitis, Kashif Rasul, Roland Vollgraf, Abdul-Saboor Sheikh, and Urs
  Bergmann. 2018.
\newblock A bandit framework for optimal selection of reinforcement learning
  agents.
\newblock In \emph{NeurIPS 2018 Workshop on Deep Reinforcement Learning}.

\bibitem[{Merity et~al.(2018)Merity, Keskar, and
  Socher}]{merity2017regularizing}
Stephen Merity, Nitish~Shirish Keskar, and Richard Socher. 2018.
\newblock Regularizing and optimizing {LSTM} language models.
\newblock In \emph{International Conference on Learning Representations}.

\bibitem[{Nallapati et~al.(2016)Nallapati, Zhou, dos Santos, Gulcehre, and
  Xiang}]{nallapati2016abstractive}
Ramesh Nallapati, Bowen Zhou, Cicero dos Santos, Caglar Gulcehre, and Bing
  Xiang. 2016.
\newblock Abstractive text summarization using sequence-to-sequence rnns and
  beyond.
\newblock In \emph{Proceedings of The 20th SIGNLL Conference on Computational
  Natural Language Learning}, pages 280--290.

\bibitem[{Nema and Khapra(2018)}]{nema2018towards}
Preksha Nema and Mitesh~M Khapra. 2018.
\newblock Towards a better metric for evaluating question generation systems.
\newblock In \emph{Proceedings of the 2018 Conference on Empirical Methods in
  Natural Language Processing}, pages 3950--3959.

\bibitem[{Neubig and Watanabe(2016)}]{neubig2016optimization}
Graham Neubig and Taro Watanabe. 2016.
\newblock Optimization for statistical machine translation: A survey.
\newblock \emph{Computational Linguistics}, 42(1):1--54.

\bibitem[{Noreen(1989)}]{noreen1989computer}
Eric~W Noreen. 1989.
\newblock \emph{Computer-intensive methods for testing hypotheses}.
\newblock Wiley New York.

\bibitem[{Papineni et~al.(2002)Papineni, Roukos, Ward, and
  Zhu}]{papineni2002bleu}
Kishore Papineni, Salim Roukos, Todd Ward, and Wei-Jing Zhu. 2002.
\newblock {BLEU}: a method for automatic evaluation of machine translation.
\newblock In \emph{Proceedings of the 40th annual meeting on association for
  computational linguistics}, pages 311--318. Association for Computational
  Linguistics.

\bibitem[{Pasunuru and Bansal(2017{\natexlab{a}})}]{pasunuru2017multi}
Ramakanth Pasunuru and Mohit Bansal. 2017{\natexlab{a}}.
\newblock Multi-task video captioning with video and entailment generation.
\newblock In \emph{Proceedings of the 55th Annual Meeting of the Association
  for Computational Linguistics (Volume 1: Long Papers)}, volume~1, pages
  1273--1283.

\bibitem[{Pasunuru and Bansal(2017{\natexlab{b}})}]{pasunuru2017reinforced}
Ramakanth Pasunuru and Mohit Bansal. 2017{\natexlab{b}}.
\newblock Reinforced video captioning with entailment rewards.
\newblock In \emph{EMNLP}.

\bibitem[{Pasunuru and Bansal(2018)}]{pasunuru2018multi}
Ramakanth Pasunuru and Mohit Bansal. 2018.
\newblock Multi-reward reinforced summarization with saliency and entailment.
\newblock In \emph{Proceedings of the 2018 Conference of the North American
  Chapter of the Association for Computational Linguistics: Human Language
  Technologies, Volume 2 (Short Papers)}, volume~2, pages 646--653.

\bibitem[{Paulus et~al.(2018)Paulus, Xiong, and Socher}]{paulus2017deep}
Romain Paulus, Caiming Xiong, and Richard Socher. 2018.
\newblock A deep reinforced model for abstractive summarization.
\newblock In \emph{International Conference on Learning Representations}.

\bibitem[{Rajpurkar et~al.(2016)Rajpurkar, Zhang, Lopyrev, and
  Liang}]{rajpurkar2016squad}
Pranav Rajpurkar, Jian Zhang, Konstantin Lopyrev, and Percy Liang. 2016.
\newblock Squad: 100,000+ questions for machine comprehension of text.
\newblock In \emph{Proceedings of the 2016 Conference on Empirical Methods in
  Natural Language Processing}, pages 2383--2392.

\bibitem[{Ranzato et~al.(2016)Ranzato, Chopra, Auli, and
  Zaremba}]{ranzato2015sequence}
Marc'Aurelio Ranzato, Sumit Chopra, Michael Auli, and Wojciech Zaremba. 2016.
\newblock Sequence level training with recurrent neural networks.
\newblock In \emph{ICLR}.

\bibitem[{Rasmussen and Williams(2005)}]{Rasmussen2015GPML}
Carl~Edward Rasmussen and Christopher K.~I. Williams. 2005.
\newblock \emph{Gaussian Processes for Machine Learning (Adaptive Computation
  and Machine Learning)}.
\newblock The MIT Press.

\bibitem[{Reiter(1995)}]{reiter1995nlg}
Ehud Reiter. 1995.
\newblock {NLG} vs. templates.
\newblock In \emph{Proc of the Fifth European Workshop on Natural-Language
  Generation}.

\bibitem[{Reiter(2007)}]{reiter2007architecture}
Ehud Reiter. 2007.
\newblock An architecture for data-to-text systems.
\newblock In \emph{Proceedings of the Eleventh European Workshop on Natural
  Language Generation}, pages 97--104. Association for Computational
  Linguistics.

\bibitem[{Reiter(2017)}]{reiter2017you}
Ehud Reiter. 2017.
\newblock You need to understand your corpora-the weathergov example.
\newblock \emph{Blogpost-https://ehudreiter. com/2017/05/09/weathergov}.

\bibitem[{Reiter and Dale(1997)}]{reiter1997building}
Ehud Reiter and Robert Dale. 1997.
\newblock Building applied natural language generation systems.
\newblock \emph{Natural Language Engineering}, 3(1):57--87.

\bibitem[{Reiter et~al.(2005)Reiter, Sripada, Hunter, Yu, and
  Davy}]{reiter2005choosing}
Ehud Reiter, Somayajulu Sripada, Jim Hunter, Jin Yu, and Ian Davy. 2005.
\newblock Choosing words in computer-generated weather forecasts.
\newblock \emph{Artificial Intelligence}, 167(1-2):137--169.

\bibitem[{Ren et~al.(2017)Ren, Wang, Zhang, Lv, and Li}]{ren2017deep}
Zhou Ren, Xiaoyu Wang, Ning Zhang, Xutao Lv, and Li-Jia Li. 2017.
\newblock Deep reinforcement learning-based image captioning with embedding
  reward.
\newblock In \emph{Proceedings of the IEEE Conference on Computer Vision and
  Pattern Recognition}, pages 290--298.

\bibitem[{Rennie et~al.(2017)Rennie, Marcheret, Mroueh, Ross, and
  Goel}]{rennie2016self}
Steven~J Rennie, Etienne Marcheret, Youssef Mroueh, Jerret Ross, and Vaibhava
  Goel. 2017.
\newblock Self-critical sequence training for image captioning.
\newblock In \emph{2017 IEEE Conference on Computer Vision and Pattern
  Recognition (CVPR)}, pages 1179--1195. IEEE.

\bibitem[{Robbins(1952)}]{robbins1952some}
Herbert Robbins. 1952.
\newblock Some aspects of the sequential design of experiments.
\newblock \emph{Bulletin of the American Mathematical Society}, 58(5):527--535.

\bibitem[{Rush et~al.(2015)Rush, Chopra, and Weston}]{rush2015neural}
Alexander~M Rush, Sumit Chopra, and Jason Weston. 2015.
\newblock A neural attention model for abstractive sentence summarization.
\newblock In \emph{CoRR}.

\bibitem[{Sankaran et~al.(2013)Sankaran, Sarkar, and Duh}]{sankaran2013multi}
Baskaran Sankaran, Anoop Sarkar, and Kevin Duh. 2013.
\newblock Multi-metric optimization using ensemble tuning.
\newblock In \emph{Proceedings of the 2013 Conference of the North American
  Chapter of the Association for Computational Linguistics: Human Language
  Technologies}, pages 947--957.

\bibitem[{See et~al.(2017)See, Liu, and Manning}]{see2017get}
Abigail See, Peter~J Liu, and Christopher~D Manning. 2017.
\newblock Get to the point: Summarization with pointer-generator networks.
\newblock In \emph{Proceedings of the 55th Annual Meeting of the Association
  for Computational Linguistics (Volume 1: Long Papers)}, volume~1, pages
  1073--1083.

\bibitem[{Serban et~al.(2016)Serban, Sordoni, Bengio, Courville, and
  Pineau}]{serban2016building}
Iulian~Vlad Serban, Alessandro Sordoni, Yoshua Bengio, Aaron~C Courville, and
  Joelle Pineau. 2016.
\newblock Building end-to-end dialogue systems using generative hierarchical
  neural network models.
\newblock In \emph{AAAI}, pages 3776--3784.

\bibitem[{Sharaf and Daum{\'e}~III(2019)}]{sharaf2019meta}
Amr Sharaf and Hal Daum{\'e}~III. 2019.
\newblock Meta-learning for contextual bandit exploration.
\newblock \emph{arXiv preprint arXiv:1901.08159}.

\bibitem[{Sharma and Ravindran(2017)}]{sharma2017online}
Sahil Sharma and Balaraman Ravindran. 2017.
\newblock Online multi-task learning using active sampling.
\newblock In \emph{ICLR 2017 Workshop}.

\bibitem[{Snover et~al.(2006)Snover, Dorr, Schwartz, Micciulla, and
  Makhoul}]{snover2006study}
Matthew Snover, Bonnie Dorr, Richard Schwartz, Linnea Micciulla, and John
  Makhoul. 2006.
\newblock A study of translation edit rate with targeted human annotation.
\newblock In \emph{Proceedings of association for machine translation in the
  Americas}.

\bibitem[{Song et~al.(2018{\natexlab{a}})Song, Wang, and
  Hamza}]{song2017unified}
Linfeng Song, Zhiguo Wang, and Wael Hamza. 2018{\natexlab{a}}.
\newblock A unified query-based generative model for question generation and
  question answering.
\newblock In \emph{NAACL}.

\bibitem[{Song et~al.(2018{\natexlab{b}})Song, Wang, Hamza, Zhang, and
  Gildea}]{song2018leveraging}
Linfeng Song, Zhiguo Wang, Wael Hamza, Yue Zhang, and Daniel Gildea.
  2018{\natexlab{b}}.
\newblock Leveraging context information for natural question generation.
\newblock In \emph{Proceedings of the 2018 Conference of the North American
  Chapter of the Association for Computational Linguistics: Human Language
  Technologies, Volume 2 (Short Papers)}, pages 569--574.

\bibitem[{Sun et~al.(2018)Sun, Liu, Lyu, He, Ma, and Wang}]{sun2018answer}
Xingwu Sun, Jing Liu, Yajuan Lyu, Wei He, Yanjun Ma, and Shi Wang. 2018.
\newblock Answer-focused and position-aware neural question generation.
\newblock In \emph{Proceedings of the 2018 Conference on Empirical Methods in
  Natural Language Processing}, pages 3930--3939.

\bibitem[{Sutskever et~al.(2014)Sutskever, Vinyals, and
  Le}]{sutskever2014sequence}
Ilya Sutskever, Oriol Vinyals, and Quoc~V Le. 2014.
\newblock Sequence to sequence learning with neural networks.
\newblock In \emph{Advances in neural information processing systems}, pages
  3104--3112.

\bibitem[{Sutton and Barto(2018)}]{sutton2018reinforcement}
Richard~S Sutton and Andrew~G Barto. 2018.
\newblock \emph{Reinforcement learning: An introduction}.
\newblock MIT press.

\bibitem[{Venugopalan et~al.(2015)Venugopalan, Rohrbach, Donahue, Mooney,
  Darrell, and Saenko}]{venugopalan2015sequence}
Subhashini Venugopalan, Marcus Rohrbach, Jeffrey Donahue, Raymond Mooney,
  Trevor Darrell, and Kate Saenko. 2015.
\newblock Sequence to sequence-video to text.
\newblock In \emph{Proceedings of the IEEE international conference on computer
  vision}, pages 4534--4542.

\bibitem[{Vinyals et~al.(2015{\natexlab{a}})Vinyals, Fortunato, and
  Jaitly}]{vinyals2015pointer}
Oriol Vinyals, Meire Fortunato, and Navdeep Jaitly. 2015{\natexlab{a}}.
\newblock Pointer networks.
\newblock In \emph{Advances in Neural Information Processing Systems}, pages
  2692--2700.

\bibitem[{Vinyals and Le(2015)}]{vinyals2015neural}
Oriol Vinyals and Quoc Le. 2015.
\newblock A neural conversational model.
\newblock In \emph{Proceedings of ICML Deep Learning Workshop}.

\bibitem[{Vinyals et~al.(2015{\natexlab{b}})Vinyals, Toshev, Bengio, and
  Erhan}]{vinyals2015show}
Oriol Vinyals, Alexander Toshev, Samy Bengio, and Dumitru Erhan.
  2015{\natexlab{b}}.
\newblock Show and tell: A neural image caption generator.
\newblock In \emph{Proceedings of the IEEE conference on computer vision and
  pattern recognition}, pages 3156--3164.

\bibitem[{Wang et~al.(2017)Wang, Yang, Wei, Chang, and Zhou}]{wang2017gated}
Wenhui Wang, Nan Yang, Furu Wei, Baobao Chang, and Ming Zhou. 2017.
\newblock Gated self-matching networks for reading comprehension and question
  answering.
\newblock In \emph{Proceedings of the 55th Annual Meeting of the Association
  for Computational Linguistics (Volume 1: Long Papers)}, pages 189--198.

\bibitem[{Wang et~al.(2018)Wang, Chen, Wu, Wang, and Yang~Wang}]{wang2018video}
Xin Wang, Wenhu Chen, Jiawei Wu, Yuan-Fang Wang, and William Yang~Wang. 2018.
\newblock Video captioning via hierarchical reinforcement learning.
\newblock In \emph{Proceedings of the IEEE Conference on Computer Vision and
  Pattern Recognition}, pages 4213--4222.

\bibitem[{Wen et~al.(2015)Wen, Gasic, Mrk{\v{s}}i{\'c}, Su, Vandyke, and
  Young}]{wen2015semantically}
Tsung-Hsien Wen, Milica Gasic, Nikola Mrk{\v{s}}i{\'c}, Pei-Hao Su, David
  Vandyke, and Steve Young. 2015.
\newblock Semantically conditioned lstm-based natural language generation for
  spoken dialogue systems.
\newblock In \emph{Proceedings of the 2015 Conference on Empirical Methods in
  Natural Language Processing}, pages 1711--1721.

\bibitem[{Williams et~al.(2018)Williams, Nangia, and
  Bowman}]{williams2018broad}
Adina Williams, Nikita Nangia, and Samuel Bowman. 2018.
\newblock A broad-coverage challenge corpus for sentence understanding through
  inference.
\newblock In \emph{Proceedings of the 2018 Conference of the North American
  Chapter of the Association for Computational Linguistics: Human Language
  Technologies, Volume 1 (Long Papers)}, pages 1112--1122.

\bibitem[{Williams(1992)}]{williams1992simple}
Ronald~J Williams. 1992.
\newblock Simple statistical gradient-following algorithms for connectionist
  reinforcement learning.
\newblock \emph{Machine learning}, 8(3-4):229--256.

\bibitem[{Wiseman et~al.(2017)Wiseman, Shieber, and
  Rush}]{wiseman2017challenges}
Sam Wiseman, Stuart Shieber, and Alexander Rush. 2017.
\newblock Challenges in data-to-document generation.
\newblock In \emph{Proceedings of the 2017 Conference on Empirical Methods in
  Natural Language Processing}, pages 2253--2263.

\bibitem[{Wu et~al.(2016)Wu, Schuster, Chen, Le, Norouzi, Macherey, Krikun,
  Cao, Gao, Macherey et~al.}]{wu2016google}
Yonghui Wu, Mike Schuster, Zhifeng Chen, Quoc~V Le, Mohammad Norouzi, Wolfgang
  Macherey, Maxim Krikun, Yuan Cao, Qin Gao, Klaus Macherey, et~al. 2016.
\newblock Google's neural machine translation system: Bridging the gap between
  human and machine translation.
\newblock \emph{arXiv preprint arXiv:1609.08144}.

\bibitem[{Xu et~al.(2015)Xu, Ba, Kiros, Cho, Courville, Salakhudinov, Zemel,
  and Bengio}]{xu2015show}
Kelvin Xu, Jimmy Ba, Ryan Kiros, Kyunghyun Cho, Aaron Courville, Ruslan
  Salakhudinov, Rich Zemel, and Yoshua Bengio. 2015.
\newblock Show, attend and tell: Neural image caption generation with visual
  attention.
\newblock In \emph{ICML}, pages 2048--2057.

\bibitem[{Yuan et~al.(2017)Yuan, Wang, Gulcehre, Sordoni, Bachman, Zhang,
  Subramanian, and Trischler}]{yuan2017machine}
Xingdi Yuan, Tong Wang, Caglar Gulcehre, Alessandro Sordoni, Philip Bachman,
  Saizheng Zhang, Sandeep Subramanian, and Adam Trischler. 2017.
\newblock Machine comprehension by text-to-text neural question generation.
\newblock In \emph{Proceedings of the 2nd Workshop on Representation Learning
  for NLP}, pages 15--25.

\bibitem[{Zaremba and Sutskever(2015)}]{zaremba2015reinforcement}
Wojciech Zaremba and Ilya Sutskever. 2015.
\newblock Reinforcement learning neural turing machines-revised.
\newblock \emph{arXiv preprint arXiv:1505.00521}.

\bibitem[{Zhang and Bansal(2019)}]{zhang2019addressing}
Shiyue Zhang and Mohit Bansal. 2019.
\newblock Addressing semantic drift in question generation for semi-supervised
  question answering.
\newblock In \emph{Proceedings of the 2019 Conference on Empirical Methods in
  Natural Language Processing and the 9th International Joint Conference on
  Natural Language Processing (EMNLP-IJCNLP)}, pages 2495--2509.

\bibitem[{Zhang and Lapata(2017)}]{zhang2017sentence}
Xingxing Zhang and Mirella Lapata. 2017.
\newblock Sentence simplification with deep reinforcement learning.
\newblock In \emph{Proceedings of the 2017 Conference on Empirical Methods in
  Natural Language Processing}, pages 584--594.

\bibitem[{Zhao et~al.(2020)Zhao, Walker, and Chaturvedi}]{zhao2020bridging}
Chao Zhao, Marilyn Walker, and Snigdha Chaturvedi. 2020.
\newblock Bridging the structural gap between encoding and decoding for
  data-to-text generation.
\newblock In \emph{ACL}.

\bibitem[{Zhao et~al.(2018)Zhao, Ni, Ding, and Ke}]{zhao2018paragraph}
Yao Zhao, Xiaochuan Ni, Yuanyuan Ding, and Qifa Ke. 2018.
\newblock Paragraph-level neural question generation with maxout pointer and
  gated self-attention networks.
\newblock In \emph{Proceedings of the 2018 Conference on Empirical Methods in
  Natural Language Processing}, pages 3901--3910.

\bibitem[{Zhou et~al.(2018)Zhou, Zhou, Corso, Socher, and Xiong}]{zhou2018end}
Luowei Zhou, Yingbo Zhou, Jason~J Corso, Richard Socher, and Caiming Xiong.
  2018.
\newblock End-to-end dense video captioning with masked transformer.
\newblock In \emph{Proceedings of the IEEE Conference on Computer Vision and
  Pattern Recognition}, pages 8739--8748.

\bibitem[{Zhou et~al.(2017)Zhou, Yang, Wei, Tan, Bao, and
  Zhou}]{zhou2017neural}
Qingyu Zhou, Nan Yang, Furu Wei, Chuanqi Tan, Hangbo Bao, and Ming Zhou. 2017.
\newblock Neural question generation from text: A preliminary study.
\newblock In \emph{National CCF Conference on Natural Language Processing and
  Chinese Computing}, pages 662--671. Springer.

\end{thebibliography}
\bibliographystyle{acl_natbib}

\appendix

\section{Exp3 Bandit Algorithm}
\label{sec:exp3-bandit-algo}
A stochastic bandit is completely determined by the distribution of rewards of respective actions. However, it will be hard to argue that rewards are truly randomly generated, and even if they are randomly generated, the rewards could be correlated over time (e.g., the validation performance at the next step will be correlated with validation performance at this time step). Taking all these factors into account makes the algorithm overly complicated, and thus an alternative is to assume nothing about the underlying mechanism that generates the rewards while still trying to achieve the lowest possible regret. This is called the adversarial bandit problem, where the goal is to design an algorithm that keeps the regret small regardless of what rewards are assigned to actions.

Exponential-weight algorithm for Exploration and Exploitation, or Exp3~\cite{auer2002nonstochastic}, was created to handle the non-stochastic adversarial bandit problem. We use this algorithm in our \fmw{} framework. Exp3 works by maintaining a set of weights for each candidate action, and the weights are used to decide randomly which action to take next. The empirical observation is fed back to the bandit to either increase or decrease the relevant weights. The algorithm also has a hyper-parameter $\gamma \in [0, 1]$ that decides the probability to take action uniformly at random.
Specifically, at round $t$, the bandit picks action (arm) $i$ among $K$ arms based on the arm selection probability which is defined as follows:

\begin{equation}
\label{eq:choosen-arm}
  p_t(i) = (1-\gamma) \frac{w_{t,i}}{\sum_{j=1}^K w_{t,j}} + \frac{\gamma}{K}
\end{equation}

where the weights $w_{t,i}$ are updated based on the observed bandit reward $r^B_t$:
\begin{equation}
    \hat{r}^B_{t,j } =
    \begin{cases*}
      r^B_t / p_t(i) & if $j = i$ \\
      0        & otherwise
    \end{cases*}
\end{equation}

\begin{equation}
\label{eq:update-bandit}
    w_{t+1,i} = w_{t,i} \exp (\gamma \hat{r}^B_{t,i} / K)
\end{equation}

\section{Training Details}
\label{sec:supp-training-details}
All the hyperparameters are tuned on the validation set for both question generation and data-to-text tasks. We use TITAN X and GeForce GTX 1080 GPUs for all our experiments, where all our RL models roughly take 1 day to train on a single GPU.  

For the question generation task, we use two layers for both bi-directional encoder and uni-directional decoder. We set the hidden size of LSTM-RNN to 600 and use BERT-based contextual embeddings as input instead of word embeddings. The number of parameters in our model is 33.3 million. We use a batch size of 32, encoder maximum length of 512 and decoder maximum length of 50, and maximum gradient clipping of~5. We use Adam optimizer~\cite{kingma2014adam} with a learning rate of 1e-3 and 1e-6 for cross-entropy model and RL models, respectively. We use a dropout of 0.3 for the cross-entropy model and no dropout for RL models. For multi-reward bandit models, we set the bandit coefficient ($\gamma$) to 0.1, and each round of the bandit consists of optimization of 100 mini-batches of training data. For HM-Bandit, we set the controller round size to 300 mini-batches. We consider the following short hyperparameters ranges
and manually tune on: learning rate in the range [1e-5, 1e-7]; bandit coefficient in the range [0.01, 0.5]; bandit round - \{10, 100\}; and controller round size - \{30, 300\}.

For WebNLG data-to-text task, we first serialize and reorder the RDF data as an intermediate planning setup, and then feed the plan into an encoder-attention-decoder style architecture with copy mechanism, to generate the text describing the RDF data. We use same hyperparameters as discussed in~\citet{zhao2020bridging} for the cross-entropy model, e.g., we use Adam with a batch size of 64, initial learning rate of 0.001, and a dropout of 0.3. All RL models are initialized with the best cross-entropy model checkpoint, and use Adam with a learning rate of 1e-6. We do not use dropout for RL models. The number of parameters in our model is 5.9 million. For multi-reward bandit models, we set the bandit coefficient ($\gamma$) to 0.15, and each round of the bandit consists of optimization of 10 mini-batches of training data. For HM-Bandit, we set the controller round size to 30 mini-batches.  We consider the following short hyperparameters ranges and manually tune on: learning rate in the range [1e-5, 1e-7]; bandit coefficient in the range [0.01, 0.5]; bandit round - \{10, 100\}; and controller round size - \{30, 300\}.

\end{document}